\documentclass[acmsmall]{acmart}

\usepackage{bm}
\usepackage{multirow}

\usepackage{soul}

\newtheorem{remark}{Remark}

\AtBeginDocument{%
  \providecommand\BibTeX{{%
    \normalfont B\kern-0.5em{\scshape i\kern-0.25em b}\kern-0.8em\TeX}}}

\setcopyright{acmcopyright}
\acmJournal{TODAES}
\acmYear{2021} \acmVolume{1} \acmNumber{1} \acmArticle{1} \acmMonth{1} \acmPrice{15.00}\acmDOI{10.1145/3495532}

\begin{document}

\title{Automatic Mapping of the Best-Suited DNN Pruning Schemes for Real-Time Mobile Acceleration}
\author{Yifan Gong}
\authornote{Both authors contributed equally to this research.}
\email{gong.yifa@northeastern.edu}
\author{Geng Yuan}
\authornotemark[1]
\affiliation{%
  \institution{Northeastern University}
  \city{Boston}
  \state{MA}
  \country{USA}}
\email{yuan.geng@northeastern.edu}

\author{Zheng Zhan}
\affiliation{%
  \institution{Northeastern University}
  \city{Boston}
  \state{MA}
  \country{USA}}
\email{zhan.zhe@northeastern.edu}

\author{Wei Niu}
\affiliation{%
  \institution{College of William and Mary}
 \city{Williamsburg}
 \state{VA}
  \country{USA}}
\email{wniu@email.wm.edu}

\author{Zhengang Li}
\affiliation{%
  \institution{Northeastern University}
  \city{Boston}
  \state{MA}
  \country{USA}}
\email{li.zhen@northeastern.edu}

\author{Pu Zhao}
\affiliation{%
  \institution{Northeastern University}
  \city{Boston}
  \state{MA}
  \country{USA}}
\email{zhao.pu@northeastern.edu}

\author{Yuxuan Cai}
\affiliation{%
  \institution{Northeastern University}
  \city{Boston}
  \state{MA}
  \country{USA}}
\email{cai.yuxu@northeastern.edu}

\author{Sijia Liu}
\affiliation{%
  \institution{Michigan State University}
  \city{East Lansing}
  \state{MI}
  \country{USA}}
\email{liusiji5@msu.edu}

\author{Bin Ren}
\affiliation{%
  \institution{College of William and Mary}
  \city{Williamsburg}
  \state{VA}
  \country{USA}}
\email{bren@cs.wm.edu}

\author{Xue Lin}
\affiliation{%
  \institution{Northeastern University}
  \city{Boston}
  \state{MA}
  \country{USA}}
\email{xue.lin@northeastern.edu}

\author{Xulong Tang}
\affiliation{%
  \institution{University of Pittsburgh}
  \city{Pittsburgh}
  \state{PA}
  \country{USA}}
\email{tax6@pitt.edu}

\author{Yanzhi Wang}
\affiliation{%
  \institution{Northeastern University}
  \city{Boston}
  \state{MA}
  \country{USA}}
\email{yanz.wang@northeastern.edu}

\renewcommand{\shortauthors}{Gong and Yuan, et al.}

\begin{abstract}
Weight pruning is an effective model compression technique to tackle the challenges of achieving real-time deep neural network (DNN) inference on mobile devices. However, prior pruning schemes have limited application scenarios due to accuracy degradation, difficulty in leveraging hardware acceleration, and/or restriction on certain types of DNN layers. 
In this paper, we propose a general, fine-grained structured pruning scheme and corresponding compiler optimizations that are applicable to any type of DNN layer while achieving high accuracy and hardware inference performance. With the flexibility of applying different pruning schemes to different layers enabled by our compiler optimizations, we further probe into the new problem of determining the best-suited pruning scheme considering the different acceleration and accuracy performance of various pruning schemes. Two pruning scheme mapping methods, one is search-based and the other is rule-based, are proposed to automatically derive the best-suited pruning regularity and block size for each layer of any given DNN. 
Experimental results demonstrate that our pruning scheme mapping methods, together with the general fine-grained structured pruning scheme, outperform the state-of-the-art DNN optimization framework with up to 2.48$\times$ and 1.73$\times$ DNN inference acceleration on CIFAR-10 and ImageNet dataset without accuracy loss.
\end{abstract}



\maketitle

\section{Introduction}
\label{sec:intro}

Model compression techniques 
have been proposed to reduce the computation and memory intensity without compromising accuracy~\cite{wen2016learning,guo2016dynamic,min20182pfpce,he2018amc,zhang2018systematic,zhang2018adam,li2019additive}. 
It is a promising solution for achieving various practical deep learning (DL)-based methods such as fingerprinting \cite{jian2021radio}, YOLO\cite{cai2021yolobile}, super-resolution \cite{zhan2021achieving}, and speech recognition \cite{dong2020rtmobile} in real time on resource-limited platforms, especially mobiles and embedded devices \cite{li2021real,zhao2020achieving,niu2020achieving}.
Among the compression techniques, weight pruning~\cite{wen2016learning,guo2016dynamic,min20182pfpce,he2018amc,he2019filter} explores and reduces the vast redundancy in the number of weights and results in structural sparsity of DNN models with fewer memory references and power consumption during inference. 

The design of a weight pruning method includes two fundamental aspects, i.e., \emph{pruning regularity} and \emph{pruning algorithm}. The former refers to the structural characteristics of the DNNs after pruning, whereas the latter determines the rule to identify the weights to be pruned. From the \emph{pruning regularity} aspect, the widely adopted pruning schemes include unstructured pruning, structured pruning, and pattern-based pruning. Specifically, unstructured pruning is flexible to prune any weights and generally yields promising accuracy. However, they are not compatible with hardware accelerations due to the irregular computation after pruning~\cite{han2015learning,guo2016dynamic,liu2018rethinking}. On the other hand, structured pruning eliminates weights while maintaining a full matrix format. It is hardware-friendly but suffers from  notable accuracy degradation due to the coarse-grained nature in pruning whole filters/channels~\cite{min20182pfpce,zhuang2018discrimination,zhu2018ijcai,ma2019tiny,zhao2019variational,Liu2020Autocompress}. Recently proposed pattern-based pruning overcomes the shortcomings of prior works by incorporating fine-grained structured pruning in a hardware-aware fashion~\cite{ma2020pconv,niu2020patdnn}, with the aid of compiler. However, pattern-based pruning is only applicable to 3$\times$3 convolutional (CONV) layers and is difficult to be generalized to fully-connected (FC) layers and CONV layers with other kernel sizes. There lacks a pruning regularity that is general and achieves high accuracy and hardware performance simultaneously.  

From the \emph{pruning algorithm} aspect, 
heuristic-based pruning was first proposed in \cite {han2015learning} and gets improvements with more sophisticated designed heuristics \cite{li2017pruning,luo2017thinet,yu2018nisp,zhuang2018discrimination,he2019filter,dong2019network}. Regularization-based pruning \cite{wen2016learning,zhang2018systematic,he2018amc,ren2019ADMMNN,li2019compressing,liu2019autocompress,yuan2019sot,yuan2019ultra,ma2019resnet,li2020ss,gong2020privacy,ma2020tiny}, on the other hand, are more mathematics-oriented.
Recent works \cite{zhang2018systematic,zhang2018adam,ren2019ADMMNN,li2019compressing,ma2020blk} achieve substantial weight reduction without hurting the accuracy by leveraging Alternating Direction Methods of Multipliers (ADMM) with dynamic regularization penalties, but these methods require the manual setting of the compression rate for each layer.

To fully exploit the potential of the pruned models on mobile devices for inference accelerations, it is necessary to incorporate compiler optimizations to support efficient sparse computation and storage. 
However, state-of-the-art compiler-based DNN execution frameworks such as TensorFlow-Lite (TFLite) \cite{TensorFlow-Lite}, Alibaba Mobile Neural Network (MNN) \cite{Ali-MNN}, and TVM \cite{chen2018tvm} do not support sparse (pruned) model inference acceleration on the mobile platforms, while the recent works PCONV \cite{ma2020pconv} and PatDNN \cite{niu2020patdnn} only have limited sparse inference support for 3$\times$3 CONV layers.

Apart from the individual limitations mentioned above, there is one additional deficiency that prevents DNN models from taking full advantage of weight pruning. Different pruning schemes result in different acceleration and accuracy performance, but prior works simply apply the same pruning scheme to the entire model, undermining the flexibility to select the best-suited pruning scheme for each layer to achieve better accuracy and acceleration performance.

This paper aims to overcome the above limitations of prior works. More specifically, we make the following {\bf contributions} towards
a general, fine-grained structured pruning scheme and two automatic pruning scheme mapping methods. 

\noindent\textbf{\textit{For the pruning scheme part:}}
\begin{itemize}
    \item We propose a novel and general pruning regularity, block-based pruning for FC layers and block-punched pruning for CONV layers with different kernel sizes, that can achieve high accuracy and high hardware inference performance simultaneously.
    \item We adopt a reweighted dynamic regularization algorithm to derive the structured sparsity with automatically determined compression rate for each layer and each block without compromising the accuracy.
    \item To extract the fine-grained structure information and exploit hardware parallelism, we propose a compiler-based mobile acceleration framework that supports the proposed pruning regularity as well as other pruning regularities. It provides the flexibility to apply different pruning schemes to different layers for a better performance of the pruned model.
\end{itemize}
\noindent\textbf{\textit{For the automatic pruning scheme mapping methods part:}}
Taking the different acceleration and accuracy performance of various pruning schemes into consideration, we probe into the new problem of determining the best-suited pruning scheme for each layer of any given DNN. We propose two automatic pruning scheme mapping methods to address this problem.
More specifically:  
\begin{itemize}
\item The first is a \textbf{search-based} method leveraging the recent concept of network architecture search (NAS) \cite{zoph2016neural,zhong2018practical,tan2019mnasnet,wu2019fbnet,cai2018proxylessnas}, which employs reinforcement learning (RL) technique to yield close-to-optimal pruning scheme mappings.
\item The second is a training-free, \textbf{rule-based} method leveraging an offline-generated latency model. It is efficient and more useful in practice.
\end{itemize}

We perform comprehensive evaluations of the proposed general pruning scheme and the two mapping methods on representative DNN models and benchmark datasets.
Experimental results demonstrate that our methods significantly outperform state-of-the-art DNN pruning framework PatDNN in terms of accuracy and latency performance. We achieve 17.22ms, 18.17ms, and 3.90ms ImageNet inference time with negligible accuracy loss on an off-the-shelf mobile phone for ResNet-50, VGG-16, and MobileNetV2, respectively. Furthermore, the search-based method only shows a slightly better performance than the rule-based method while the rule-based method is training-free in pruning scheme mapping.

\section{Background and Related Works}
\label{sec:background}

\subsection{DNN Pruning: Regularity and Algorithm} \label{sec:weight_pruning}

\subsubsection{\textbf{Pruning Regularity}} 
\begin{figure*}[t]
    \centering
    \includegraphics[width=1 \textwidth]{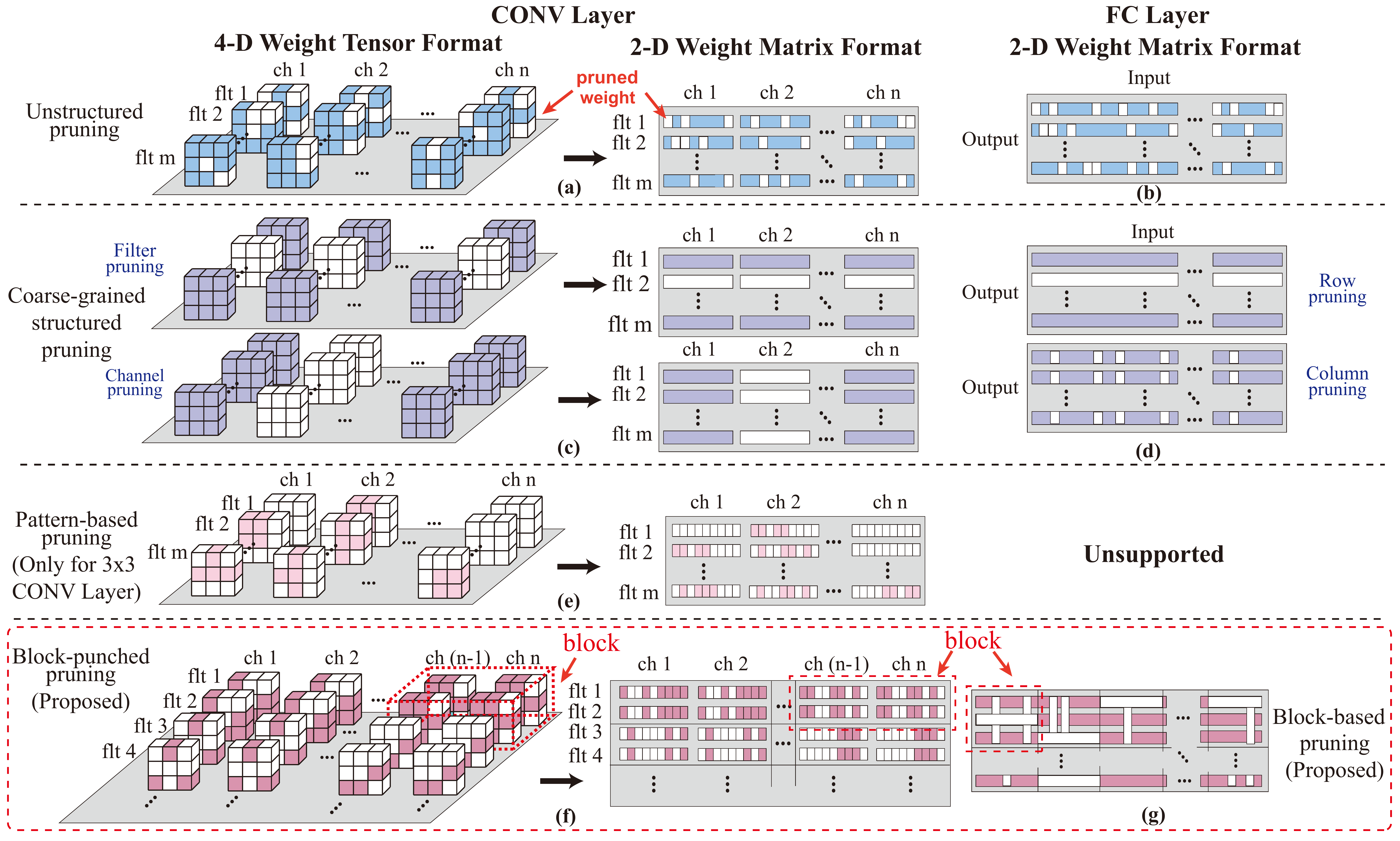}
    \caption{{Different weight pruning schemes for CONV and FC layers using 4D tensor and 2D matrix representation.}}
    \label{fig:prune_type}
\end{figure*}

From the pruning regularity aspect, existing pruning schemes can be divided into three categories: fine-grained \textit{unstructured pruning}, coarse-grained \textit{structured pruning}, and \textit{pattern-based pruning}.  We show the different pruning regularities in Fig.~\ref{fig:prune_type}, with colored grids representing remaining weights. The left and middle column in the figure illustrates pruning regularities in the 4-D weight tensor format and 2-D weight matrix format for CONV layers, respectively. The right column shows the different regularities for FC layers.

\textbf{Unstructured pruning} is fine-grained and flexible in removing weights at arbitrary locations~\cite{guo2016dynamic,frankle2018lottery,dai2019nest,dai2019grow}, as shown in Fig. \ref{fig:prune_type} (a) and (b). Though having the advantage in maintaining accuracy, unstructured pruning leads to sparse and irregular weight matrices, and as a result, indices are required to locate the non-zero weights in the sparse matrix storage format, e.g., CSR format. Therefore, it cannot effectively and efficiently leverage the hardware parallelism provided by the underlying system. Consequently, unstructured pruning is generally not compatible with GPU acceleration for DNN inference, and speed degradation can often be observed~\cite{ma2019nonstructured}.

\textbf{Structured pruning}~\cite{wen2016learning,he2017channel,he2019filter,yu2018nisp,yuan2021forms} 
focuses on CONV layers and maintains structured regularity. It consists of filter pruning and channel pruning that prune the entire filter(s)/channel(s). In the weight matrix format representation as shown in Fig.~\ref{fig:prune_type} (c), filter pruning corresponds to reducing one row of the weight matrix and it is also termed as row pruning. Accordingly, channel pruning corresponds to reducing multiple consecutive columns. The key advantage of structured pruning is that a full matrix will be maintained with dimension reduction, thereby facilitating hardware acceleration. 
However, structured pruning is coarse-grained and often leads to certain accuracy degradation \cite{Wang2019NonstructuredDW,niu2020patdnn}. 

\textbf{Pattern-based pruning}~\cite{niu2020patdnn,ma2020pconv,yuan2021mest} alleviates the shortcomings of prior works by incorporating the benefits from fine-grained pruning while maintaining structures that can be exploited for hardware accelerations with the help of compiler. Pattern-based pruning is a combination of kernel pattern pruning and connectivity pruning as shown in Fig. \ref{fig:prune_type} (e). Kernel pattern pruning prunes weights at an intra-kernel level by enforcing the locations of the remaining weights in a kernel to form a specific kernel pattern. Different kernels can have different kernel patterns, but the total types of kernel patterns are restricted to a fixed-size set.
Each kernel pattern reserves 4 non-zero weights to match the single-instruction multiple-data (SIMD) architecture of embedded CPU/GPU processors to maximize the hardware throughput. 
As a {\it fixed} number of weights are pruned, the compression rate is constant for kernel pattern pruning. 
For a higher compression rate, connectivity pruning is adopted as the supplementary to kernel pattern pruning. Connectivity pruning prunes weights at an inter-kernel level via cutting the connections between certain input and output channels. 

However, pattern-based pruning is designed for 3$\times$3 CONV layers and suffers difficulty when generalized to CONV layers with other kernel sizes and FC layers.
To avoid the execution overhead of branching conditions caused by using different pattern types, pattern-based pruning requires limiting the maximum number of different pattern types to be used. Generally, 8 or 16 different pattern types are allowed to be selected from all possible 4-entry pattern combinations to ensure a decent acceleration while not hurting accuracy.
For larger kernel sizes such as 5x5/7x7, 4-entry patterns need to be selected from 25/49 weights (instead of 9 weights in 3x3 case), making the pattern have too many potential candidates. As a result, if only 8 or 16 patterns are used, there will be an accuracy degradation.
Moreover, as studied in~\cite{ma2020image}, the Gaussian filter-like patterns and the Enhanced Laplacian of Gaussian (ELoG) filter-like patterns (as shown in Fig.~\ref{fig:prune_type} (e)) are more preferred since they can provide an enhancement on feature extraction. But such 4-entry patterns in 5x5/7x7 kernels cannot provide the receptive field size that the large kernels are supposed to have.
For the 1x1 CONV layer, there is only one weight in a kernel, making the pattern-based 
pruning same as unstructured pruning, which is hard to achieve actual acceleration.
Therefore, the existing pattern-based pruning is only suitable for 3x3 kernels, which significantly restricts the application scenarios of pattern-based pruning.

\subsubsection{\textbf{Pruning Algorithm}} There are two main categories of pruning algorithm, i.e., heuristic-based algorithm and regularization-based algorithm. Heuristic-based pruning algorithm was first proposed to achieve unstructured pruning by pruning weights with small magnitudes in an iterative manner \cite{han2015learning}. Later heuristic works get improved in multiple directions including structured-preserving pruning \cite{li2017pruning,luo2017thinet,yu2018nisp}, combining growth of neurons and connections with pruning \cite{dai2019nest}, and 
introducing
meticulously-designed criteria \cite{yu2018nisp,zhuang2018discrimination,he2019filter,luo2017thinet} to replace magnitudes for the pruning. 

Regularization-based algorithm deal with the pruning problem using a more mathematics-oriented method. To solve filter/channel pruning problems, early works \cite{wen2016learning,he2017channel} incorporate $\ell_1$ or $\ell_2$ structured regularization in the loss function. Work \cite{liu2017learning} introduces a scaling factor to each channel while imposing $\ell_1$ regularization on the scaling factors in batch normalization to prune channels with near-zero scaling factors. However, these works directly apply fixed regularization terms that penalize all weights equally, incurring potential accuracy loss. Later works \cite{zhang2018systematic,ren2019ADMMNN,gong2020privacy} adopt ADMM to reform the pruning problem as optimization problems with dynamic regularization penalties, thus preserving accuracy. One drawback of these methods is the requirement for the manual setting of the compression rate for each layer.

\subsection{Compiler-based DNN Frameworks on Mobile}

Mobile devices become key carriers of deep learning~\cite{hegde2016caffepresso, lane2017squeezing, ota2017deep, zhang2019deep} to enable the widespread of machine intelligence. 
To facilitate the deployment of various DNN models on mobile devices, multiple mobile DNN execution frameworks from both industry and academia attract broad attention \cite{lane2016deepx,lane2015deepear,xu2018deepcache,huynh2017deepmon,yao2017deepsense,han2016mcdnn}.
TFLite~\cite{TensorFlow-Lite}, MNN~\cite{Ali-MNN}, and TVM~\cite{chen2018tvm} are three representative state-of-the-art end-to-end DNN execution frameworks with high execution efficiency. 
They employ several performance optimization techniques, such as various computation graph optimizations, tensor optimizations, and half-float support. Particularly, TVM includes a more advanced parameter auto-tuning technique.
However, none of these frameworks support sparse (pruned) DNN models on mobile platforms\footnote{TVM considers sparsity recently for desktop processors.}. This is the essential drawback that obstructs the real-time DNN inference on mobile devices.
Take VGG-16 network, one of the key DNN models in transfer learning, as an example, TVM takes 200ms to perform an inference on the embedded  GPU  (Adreno  640), and TFLite takes an even longer time (270 ms).

Previous efforts based on fine-grained pattern-based pruning such as PatDNN~\cite{niu2020patdnn} and PCONV~\cite{ma2020pconv} employ a set of compiler-based optimizations to support sparse DNN models, significantly accelerating the end-to-end DNN inference on mobile devices. 
However, they mainly accelerate the square and small convolution kernels used in 3$\times$3 CONV layers. A larger kernel size, e.g., 5$\times$5, 7$\times$7, will introduce huge code execution overhead due to the increasing number of branches in generated code. In addition, they cannot support FC layers and 1$\times$1 CONV layers that are commonly used in DNNs.

\section{Overview of the Automatic Pruning Scheme Mapping Framework}
\begin{figure} [t]
     \centering
     \includegraphics[width=0.65\columnwidth]{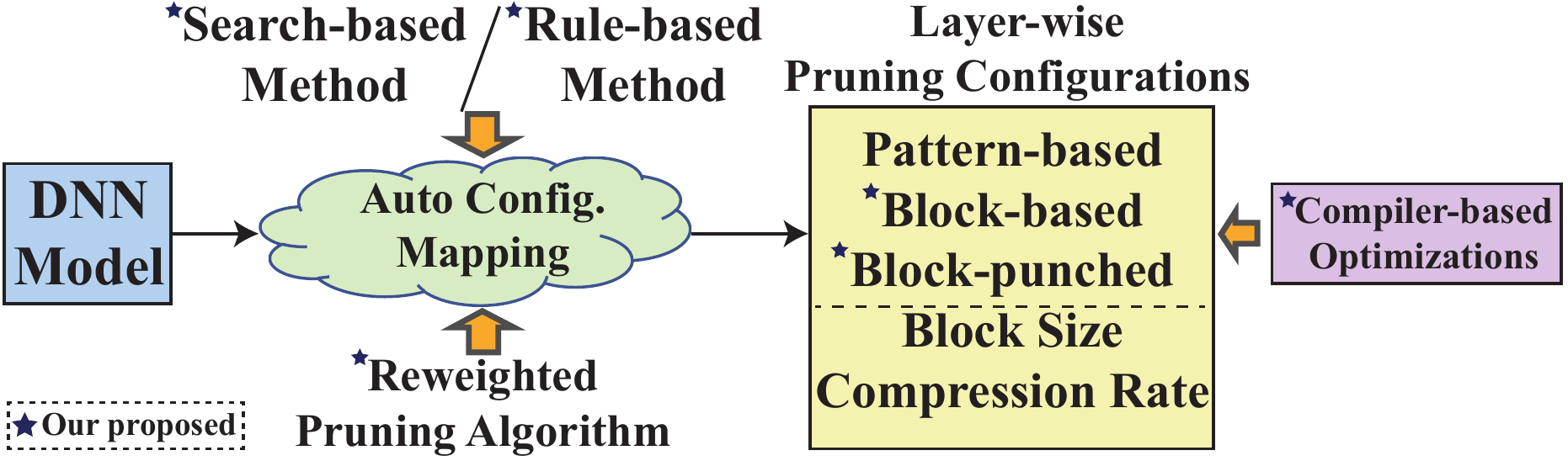}     
     \caption{High-level overview of proposed automatic pruning scheme mapping framework.}
     \label{fig:high_level_overview}
\end{figure}
To achieve real-time mobile acceleration for various modern DNNs, we propose an automatic pruning scheme mapping framework, which is illustrated in Fig. \ref{fig:high_level_overview}. Given an arbitrary DNN model, the framework can automatically map the best pruning configurations to each layer and leverage compiler-based optimizations to achieve inference speedup. The layer-wise configurations include the pruning regularity, compression rate, and the block size.

In order to achieve the design objective, our framework contains the following innovations. We first propose a general, fine-grained pruning regularity that is applicable to \textbf{different} types of layers while achieving both high accuracy and hardware acceleration performance to overcome the limitations of prior pruning regularities in Section \ref{sec:pruning_regularity}. To determine the compression rate for each layer automatically without compromising accuracy, we introduce a reweighted pruning algorithm in Section \ref{sec:reweighted}. For the 
goal of transforming compression to real inference speedup on mobile devices, we propose corresponding compiler-based optimizations that support the proposed pruning regularity as well as other pruning regularities in Section \ref{sec:compiler_opt}. As directly applying the same pruning scheme to the entire model can not yield the optimal performance, we further propose to map the best-suited pruning configurations to each layer of any given DNN for mobile devices thanks to the flexibility enabled by our compiler optimizations. The mapping methods include a comprehensive search-based method that can provide close-to-optimal results in Section \ref{sec:search_based_method} and a training-free rule-based method that is more useful in-practice while reaching similar performance as the search-based method in Section \ref{sec:rule_based_method}.

\section{General, Fine-grained Structured Pruning Scheme}
\label{sec:block_pruning}

In this section, we present a novel fine-grained structured pruning scheme and corresponding compiler optimizations to (i) achieve high accuracy and hardware inference performance simultaneously while applicable to different types of layers; (ii) determine the compression rate for each layer automatically without compromising the accuracy; and (iii) provide the supports to the proposed pruning regularity and other pruning regularities for the exploitation of the hardware parallelism.
We start by providing a general fine-grained structured pruning regularity 
that includes block-based pruning for FC layers and block-punched pruning for CONV layers with different kernel sizes in Section \ref{sec:pruning_regularity}. Next, a reweighted dynamic regularization algorithm which allows the automatic determination of the per-layer and per-block compression rate is introduced to derive the sparse regularity in Section \ref{sec:reweighted}. Then we provide corresponding compiler optimizations for the proposed pruning scheme to enable efficient on-device inference of the pruned model in Section \ref{sec:compiler_opt}. 

\subsection{General, Fine-Grained Structured Pruning Regularity} \label{sec:pruning_regularity}
\begin{figure} [t]
     \centering
     \includegraphics[width=0.75\columnwidth]{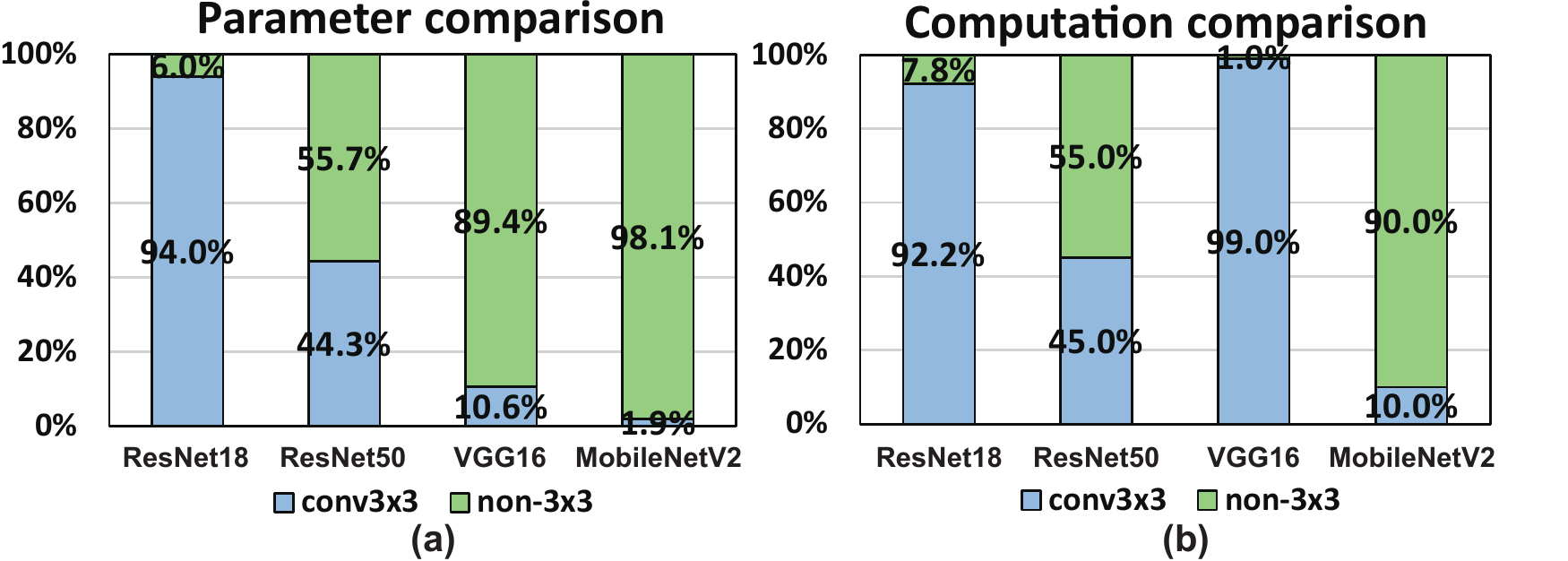}     
     \caption{Comparisons of (a) parameter ratio and (b) computation ratio for 3$\times$3 CONV layers and non-3$\times$3 layers for different networks on ImageNet dataset.}
     \label{fig:para_ratio}
\end{figure}

Though state-of-the-art pattern-based pruning strikes a desirable balance between accuracy and hardware efficiency, it \textbf{only} works for CONV layers with $3\times3$ kernels and \textbf{suffers difficulty} when generalized to layers with other kernel sizes and FC layers. Note that not all of the layers only operate on $3\times3$ kernels in a given DNN model. As a result, the number of layers using $3\times3$ kernels affect the effectiveness of pattern-based pruning. Fig.~\ref{fig:para_ratio} illustrates the percentage of the parameters and multiply-and-accumulates (MACs) in $3\times3$ CONV layers of four representative networks. The large portion of non-$3\times3$ CONV layers leaves great space for higher compression rate and faster inference that cannot be achieved by pattern-based pruning alone.

To alleviate the deficiencies,
we propose a \textbf{general} pruning scheme with fine-grained structured pruning regularity, including block-based pruning for FC layers and block-punched pruning for CONV layers with different kernel sizes. 

\subsubsection{\textbf{Block-based Pruning for FC Layers}}
Block-based pruning is an extension of the coarse-grained structured pruning that prunes rows/columns in matrix-based computation for FC layers. 
As shown in Fig.~\ref{fig:prune_type} (g), we divide a whole weight matrix of a FC layer to a number of equal-sized blocks (e.g., 4$\times$4, 16$\times$32, 64$\times$128, etc.), and apply independent row and column pruning for each block. 
The compression rate (the number of pruned rows/columns) for each block can either be the same or different, which depends on the design requirements.

\subsubsection{\textbf{Block-punched Pruning for CONV Layers}}
Compared with matrix-based representation and computation, tensor-based representation and computation are more suitable for CONV layers. Thus, inspired by block-based pruning, we further propose block-punched pruning that is tailored for CONV layers and can be accelerated with the same compiler optimizations. As shown in Fig. \ref{fig:prune_type} (f), block-punched pruning first partitions the weight tensor of a CONV layer into groups (blocks) of kernels along the filter and input channel dimensions. For each block, the weights at the same locations for all kernels within the block are pruned. With effective compiler-level executable code generation, high hardware parallelism and inference acceleration on mobile can be achieved. 

Compared with state-of-the-art pattern-based pruning, the proposed fine-grained structured pruning regularity is \textbf{general and flexible} as it can adaptively prune FC layers and CONV layers with different kernel sizes. In the same time, block-based pruning and block-punched pruning can simultaneously achieve \textbf{high accuracy} and \textbf{high hardware inference performance} like pattern-based pruning. 
The \textbf{high accuracy} is attributed to the fine-grained property of pruning regularity, which allows higher flexibility when searching the pruned model structure compared to coarse-grained structured pruning that prunes entire rows/columns in weight matrices. 
On the other hand, the \textbf{high hardware inference performance} is attributed to the appropriate degree of structural regularity, which can be exploited by compiler-level code generation to achieve high or even maximum hardware parallelism. 
With an appropriate selection of the block size, the remaining entries in each block can still be sufficient to exploit high hardware parallelism. The block size for each layer is an important hyperparameter that influences hardware performance and accuracy. We will elaborate on how to select the appropriate block size for each layer in Section \ref{sec:block_size_selection}.

\subsection{Reweighted Dynamic Regularization Algorithm} \label{sec:reweighted}

Another important design aspect of a pruning scheme is the pruning algorithm. Prior pruning algorithms such as using the group Lasso regularization~\cite{wen2016learning,he2017channel,liu2017learning} or ADMM~\cite{zhang2018systematic,ren2019admm,li2019compressing}, either suffer from potential accuracy loss or require compression rate tuning manually.
To overcome the limitations, we propose to adopt reweighted group Lasso~\cite{candes2008enhancing} method to discover the structured sparsity with systematically and dynamically reweighted penalties. More specifically, the reweighted method reduces the penalties on weights with larger magnitudes, which are likely to be more critical weights, and increases the penalties on weights with smaller magnitudes. 
A comparison of the characteristics of different regularization-based pruning algorithms is shown in Table~\ref{tab:algorithms}.  

\begin{table} [t]
     \centering
     \caption{Comparison of different pruning algorithms.}
     \includegraphics[width=0.5\columnwidth]{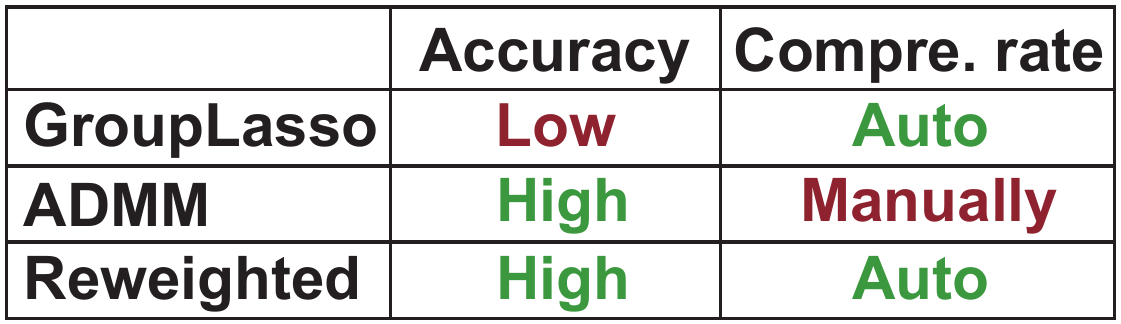}     
     \label{tab:algorithms}
\end{table}

For the $i$-th layer in the DNN, if the layer is a FC layer, let $\bm{W}_i \in \mathbb{R}^{P_i\times Q_i}$ denote the 2-D weight matrix, with $P_i$ and $Q_i$ indicating the rows and columns of the weight matrix, otherwise $\bm{W}_i \in \mathbb{R}^{P_i\times Q_i\times K_i^h \times K_i^w}$ represents the 4-D weight tensor of a CONV layer, where $P_i$ is the number of filters, $Q_i$ is the number of input channels, $K_i^w$ and $K_i^h$ are the kernel width and kernel height. Let $\bm{b}_i \in \mathbb{R}^{P_i}$ represent the bias for the $i$-th layer. We also define $\bm{W} := \{\bm{W}_i\}_{i=1}^N$, and $\bm{b} := \{\bm{b}_i\}_{i=1}^N$ as the set of all weights and biases in the DNN. We denote the loss of the DNN under dataset $\mathcal{D}$ by $f(\bm{W},\bm{b};\mathcal{D})$. Each $\bm{W}_i$ is divided into $J$ blocks with the same size, $p_i\times q_i$ for a FC layer and $p_i \times q_i \times K_i^h \times K_i^w$ for a CONV layer, namely, $\bm W_i = [\bm W_{i1},  \bm W_{i2},...,\bm W_{iJ}]$, where $\bm W_{ij} \in \mathbb R^{p_i \times q_i}$ for a FC layer and $\bm W_{ij} \in \mathbb R^{p_i \times q_i \times K_i^h \times K_i^w}$ for a CONV layer. The general reweighted pruning problem is formulated as 
\begin{equation} \label{prob_block}
\small
 \underset{\bm{W},\bm{b}}{\text{minimize\quad}}
 f \big( \bm{W},\bm{b};\mathcal{D} \big)+\lambda\sum_{i=1}^{N} R(\bm{\alpha}_{i}^{(t)},\bm{W}_{i}), \\
\end{equation}
where $\lambda$ is the hyperparameter to adjust the relative importance between accuracy and sparsity. Let $\bm{\alpha}_{i}^{(t)}$ denote the collection of penalty values applied on the weights $\bm{W}_i$ for layer $i$ at step $t$. Note that each element in $\bm{\alpha}_{i}^{(t)}$ is a positive value that is determined by reweighted $\ell_{1}$ algorithm \cite{candes2008enhancing}.

For block-based row pruning, the regularization term is
\begin{equation}
\small
R({\bm{\alpha}}_{i}^{(t)},\bm{W}_{i}) = \sum_{j=1}^{J}\sum_{m=1}^{p_i} \big\| \alpha_{ijm}^{(t)} \circ  [\bm W_{ij}]_{m,:} \big\|_F^2, \label{eq:row}
\end{equation}
where the operator $\circ$ represents element-wise multiplication, $[\bm W_{ij}]_{m,:}$ denotes the $m$-th row of $\bm W_{ij}$, and $\alpha^{(t)}_{ijm}$ is updated by $\alpha^{(t)}_{ijm} = \frac{1}{\|[\bm W_{ij}]^{t}_{m,:}\|_F^2 + \epsilon}$ to help increase the degree of sparsity beyond group Lasso regularization. 

For block-based column pruning, the regularization term is
\begin{equation}
\small
R({\bm{\alpha}}_{i}^{(t)},\bm{W}_{i}) = \sum_{j=1}^{J}\sum_{n=1}^{q_i} \big\| \alpha_{ijn}^{(t)} \circ  [\bm W_{ij}]_{:,n} \big\|_F^2,
\label{eq:column}
\end{equation}
where $[\bm W_{ij}]_{:,n}$ is the $n$-th column of $\bm W_{ij}$ and $\alpha^{(t)}_{ijn}$ is updated by $\alpha^{(t)}_{ijn} = \frac{1}{\|[\bm W_{ij}]^{t}_{:,n}\|_F^2 + \epsilon}$.
The block-based row pruning problem \eqref{eq:row} and column pruning problem \eqref{eq:column} can be solved separately or simultaneously using a standard deep learning solver. 

For block-punched pruning, the regularization term is formulated as
\begin{equation}
\small
R({\bm{\alpha}}_{i}^{(t)},\bm{W}_{i}) = \sum_{j=1}^{J}\sum_{m=1}^{K_i^h}\sum_{n=1}^{K_i^w} \big\| \alpha_{ijmn}^{(t)} \circ  [\bm W_{ij}]_{:,:,m,n} \big\|_F^2, \label{eq:punched}
\end{equation}
where $[\bm W_{ij}]_{:,:,m,n}$ indicates the weight located at the $m$-th row and $n$-th column in a kernel for all kernels in the block and $\alpha^{(t)}_{ijmn} = \frac{1}{\|[\bm W_{ij}]^{t}_{:,:,m,n}\|_F^2 + \epsilon}$. 
The reweighted method only requires the hyperparameter $\lambda$ and the soft constraints formulation allows the automatic determination of the compression rate for each layer and each block.

\subsection{Compiler Optimizations for Proposed Pruning Regularity} \label{sec:compiler_opt}

\begin{figure} [t]
     \centering
     \includegraphics[width=0.65\columnwidth]{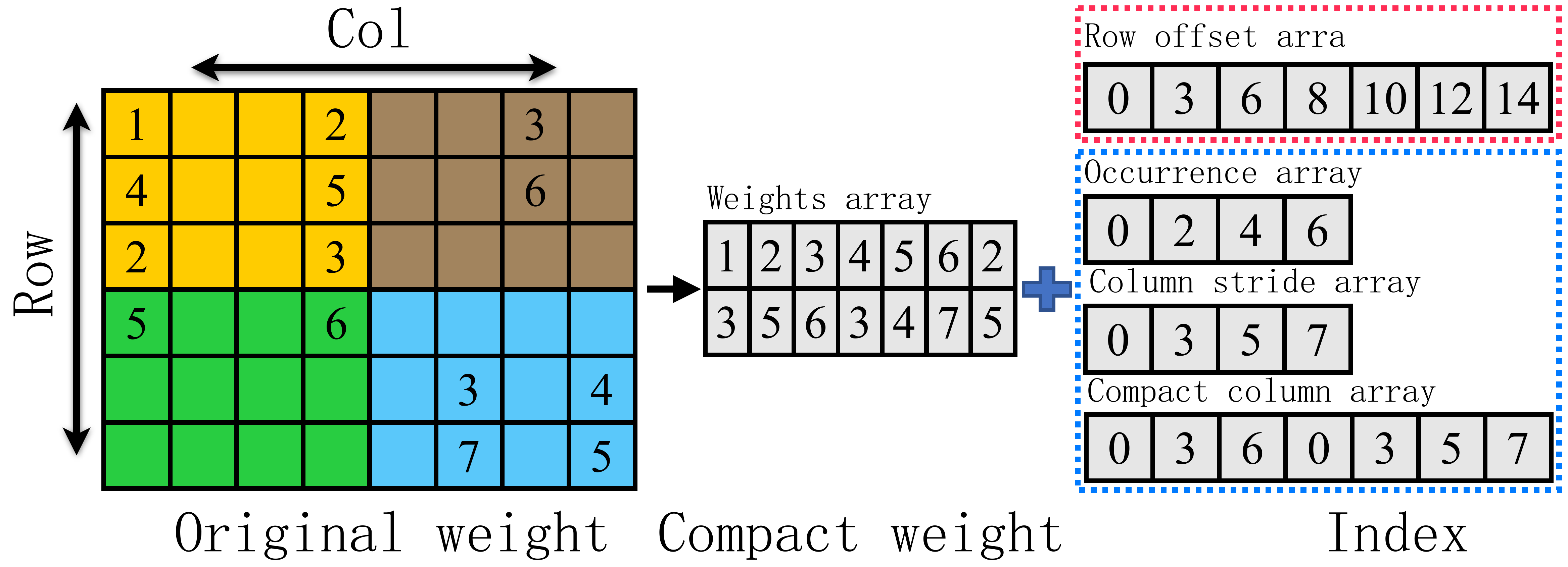} 
     \caption{Blocked Compressed Storage (BCS) for weights.}
     \label{fig:compiler_block}
\end{figure}
Compiler optimizations can turn the sparsity
of pruned models into higher speedups. Without compiler optimizations, the
pruned weights (with zero values) still participate in the inference computations, resulting in minor inference speedup. Hence, we develop a comprehensive compiler-based automatic code generation framework to extract the fine-grained structure information in block-punched and block-based pruning. The framework also supports other pruning regularities including unstructured pruning, structured pruning, and pattern-based pruning.
Our proposed compiler-based mobile acceleration framework first compacts the model storage with a novel compression format called Blocked Compressed Storage (\textbf{BCS}) format, as shown in Fig.~\ref{fig:compiler_block}. Then, it performs computation reordering to reduce the branches within each thread and eliminate the load imbalance among threads.

\textbf{BCS} stores non-zero weights as Compressed Sparse Row format (CSR) with a better compression rate by further compressing the index with a hierarchical structure. Traditional CSR has to store each non-zero weight with an explicit column index. Our proposed block-based/block-punched pruning preserves non-zero weights in identical columns within each block, inducing many repeated column indices if we use CSR. BCS eliminates this redundancy with a hierarchical compression on the column index only. 

Fig.~\ref{fig:compiler_block} shows a simplified example. {\tt Weights array} stores all non-zero weights. {\tt Compact column array} stores the compressed column index, e.g., [0, 3, 6] denotes the column id of the first three weights [1, 2, 3]. {\tt Column stride array} denotes the start and end index of each row in compact column array, e.g., [0, 3] denotes that the column index for the first row starts from index 0 and ends at index 2 in compact column array. The same column indices may be used for multiple rows. {\tt Occurrence array} is used to specify the start and end rows with the identical column index, e.g., [0, 2] means that row 0 and 1 share the same column index. BCS also contains a {\tt row offset array} to specify the starting location of each row in weight array.

Usually, the weight distribution is not as regular as the above simplified example, thus, a row reordering optimization is also included to further improve the regularity of the weight matrix. After this reordering, the continuous rows with identical or similar numbers of non-zero weights are processed by multi-threads simultaneously, thereby eliminating thread divergence and achieving load balance. Each thread processes more than one rows, thus eliminating branches and improving instruction-level parallelism. We also incorporate other compiler-based optimizations for on-mobile DNN inference acceleration, such as the layer-fusion, the auto-tuning, and the high-level domain-specific language. More details are provided in the Appendix.

\subsection{Effectiveness of the Proposed Pruning Scheme}

\begin{figure} [t!]
     \centering
     \includegraphics[width=0.6\columnwidth]{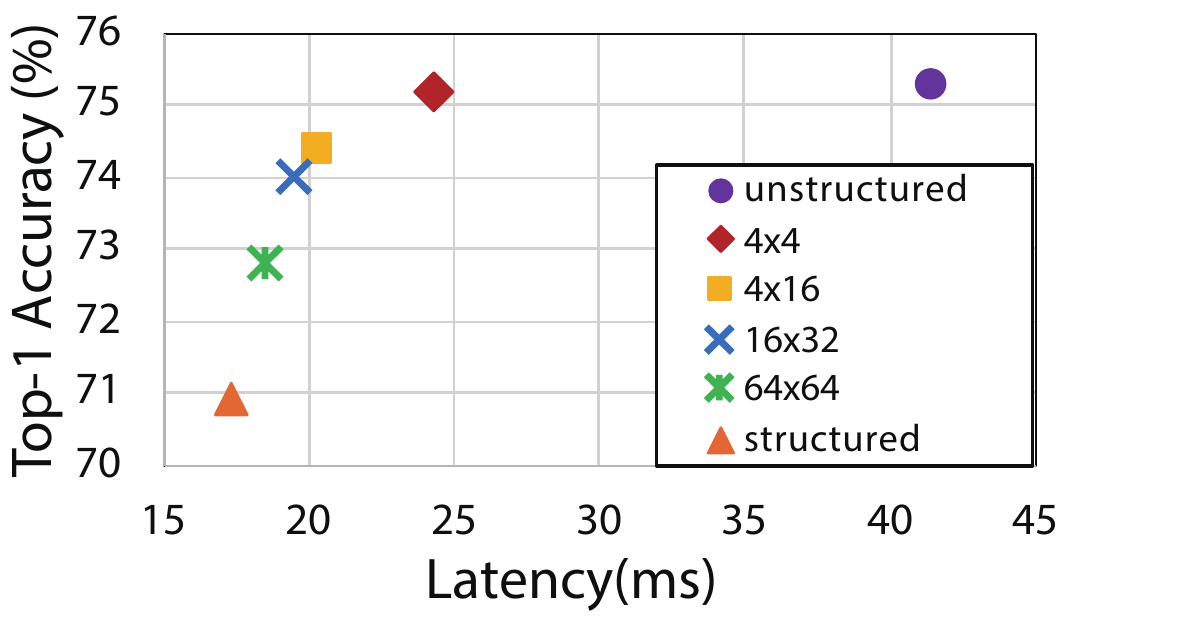}
     \caption{Accuracy and latency performance with different block sizes on ResNet-50 using ImageNet dataset.}
     \label{fig:sample_block_result}
\end{figure}

We show an example of the inference accuracy and acceleration performance of the proposed pruning scheme on ResNet-50 using ImageNet dataset in Fig. \ref{fig:sample_block_result}. More thorough evaluation results are presented in Section \ref{sec:block_related_results}. Here, block-based pruning is applied for \textit{all} FC layers and block-punched pruning is applied for \textit{all} CONV layers. The compression rate for each layer is derived by the reweighted dynamic regularization algorithm. As can be seen from the figure, unstructured pruning, which is equivalent as setting the block size as $1\times1$ for each layer, achieves the highest accuracy while has the worst performance in latency. 
In contrast, structured pruning, i.e., using the whole matrix as the block size, achieves the fastest inference but degrades the accuracy the most. With a suitable block size, our proposed fine-grained structured pruning scheme achieves high accuracy and inference speed simultaneously. The reason is that the maximal hardware parallelism is limited by the computation resource. Since the weight matrix/tensor is typically very large, the remaining entries in each block is still sufficient to exploit high hardware parallelism.
With parallelism maximally exploited, the hardware inference performance can be almost the same as coarse-grained structured pruning.      

\textbf{Takeaway:} In this section, we first introduced a \textbf{general}, fine-grained structured pruning regularity, which can work for CONV layers with any kernel size and FC layers. Second, we proposed reweighted group Lasso with block-based constraints as the pruning algorithm to derive the structured sparsity with \textbf{automatically} determined compression rate for each layer and each block. Third, we develop the first compiler-based mobile acceleration framework that supports general block-based/block-punched sparsity as well as other pruning regularities, which is flexible and allows
different layers to adopt different pruning regularities and block sizes.

\section{Automatic Pruning Scheme Mapping Methods for Mobile Devices}
\label{sec:framework}

Though the general, fine-grained pruning scheme proposed in Section \ref{sec:block_pruning} can achieve high accuracy and hardware acceleration performance, it is not optimal to directly apply the same pruning scheme to the entire model as different layers may prefer different pruning regularities and configurations, e.g., the compression rate and block size.
Fortunately, effective compiler optimization techniques provide the flexibility to apply different pruning regularities and block sizes to different layers. 
As different weight pruning schemes have different acceleration and accuracy performance under the same mobile acceleration framework, it is important to have a pruning scheme mapping method to determine the pruning configurations for each layer. Therefore, we further probe into the problem of mapping the best-suited pruning scheme for each layer of any given DNN to obtain pruned model with better performance in terms of accuracy and latency in this section. 

\begin{figure*} [t]
     \centering
     \includegraphics[width=1\textwidth]{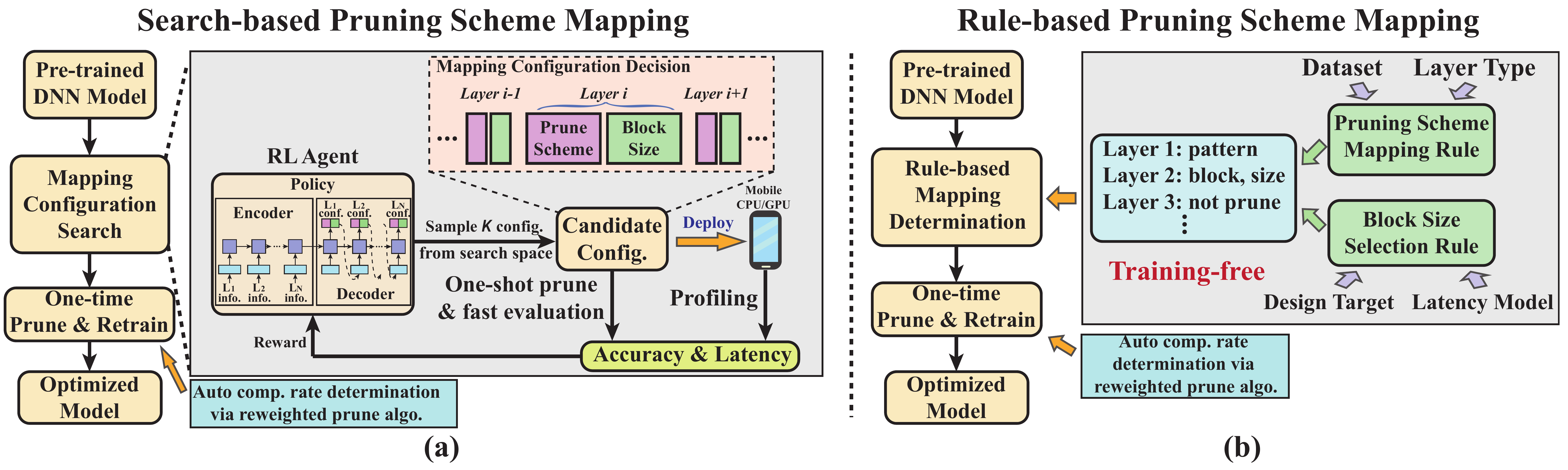}  
     \caption{Overview of (a) Search-based pruning scheme mapping, and (b) rule-based pruning scheme mapping.}
     \label{fig:search_vs_rule}
\end{figure*}

The performance of a pruned model is influenced by the compression rate, pruning regularity, and block size when block-based/block-punched pruning is selected, of each layer. This is a new challenge resulted from the new dimension of compiler-aware pruning scheme optimizations. To find the appropriate pruning schemes in such a large design space, we propose two automatic pruning scheme mapping methods, one is \textbf{search-based} and the other is \textbf{rule-based}, as shown in Fig. \ref{fig:search_vs_rule}. The former is a more comprehensive framework to yield close-to-optimal pruning scheme mapping results, while the latter is a \emph{training-free} procedure that is efficient and more useful in practice. Note that with our proposed reweighted dynamic regularization algorithm in Section \ref{sec:reweighted}, the compression rate can be obtained automatically for each layer and each block. Thus, the search space of the pruning scheme mapping problem can be reduced to finding the appropriate pruning regularity and the block size for each layer in the given DNN.

\subsection{Search-based Pruning Scheme Mapping Method} \label{sec:search_based_method}
Albeit we simplify the search space with the reweighted dynamic regularization algorithm to determine the per-layer and per-block compression rate automatically, there is still a huge amount of combinations of pruning regularities and block sizes to seek. Inspired by recent advances in NAS \cite{zoph2016neural,zhong2018practical,tan2019mnasnet,wu2019fbnet,cai2018proxylessnas}, we consider to leverage a search-based method by employing RL \cite{li2016learning,sutton2018reinforcement} to map the appropriate pruning scheme for each layer of a given DNN. 

In RL, an agent interacts with the environment by taking an action $a_t \in A$ according to a policy $\pi$ upon the observation of a state $s_t \in S$ at time step $t$. For our problem, each time step $t$ corresponds to the pruning scheme mapping of one layer. The state $s_t \in S$ represents the information of current layer, which is defined as a $4$-D vector \{\textit{layer type}, \textit{kernel size}, \textit{input channel number}, \textit{output channel number}\}. The action $a_t \in A$ is the mapping decision for the current layer, which is a $2$-D vector \{\textit{pruning regularity}, \textit{block size}\}. For a $N$-layer DNN with information $\mathcal{I} = \{s_1,\cdots,s_N\}$, an entire mapping $\mathcal{M} = \{a_1,\cdots,a_N\}$ can be found with $N$ time steps. Let $R(\mathcal{M})$ denote the cumulative reward for $\mathcal{M}$, which is the optimization target of the RL agent. A good pruning scheme mapping should achieve high accuracy and hardware performance jointly, thus we define $R(\mathcal{M})$ as the weighted sum of the accuracy and the negative of the latency of the pruned model with information $\mathcal{I}$ under the mapping $\mathcal{M}$.

We leverage the policy gradient method \cite{sutton1999policy} to directly learn a parameterized policy for the pruning scheme mapping, and the training objective of the policy is defined as:
\begin{equation}
\small
J(\theta) = \mathbb{E}_{\mathcal{M}\sim \pi(\mathcal{M}|\mathcal{I};\theta)}[R(\mathcal{M})|\mathcal{I}],
\end{equation}
where $\pi(\mathcal{M}|\mathcal{I};\theta)$ is a sequence-to-sequence model in our work. The input to the encoder RNN is the sequence of the information of each layer in the target DNN and the decoder is an LSTM with $N$ time steps to output the mapping decision for each layer at the same encoder time step.
We estimate the gradient of the objective function by drawing $K$ mapping decision samples from $\mathcal{M}_k\sim \pi(\mathcal{M}|\mathcal{I};\theta)$ and reduce the variance of the estimate with a baseline term $B$, leading to
\begin{equation}
\small
    \nabla_\theta J(\theta) \approx \frac{1}{K}\sum_{k=1}^K (R(\mathcal{M}_k)-B)\cdot\nabla_\theta \log\pi(\mathcal{M}_k|\mathcal{I};\theta).
\end{equation}
For each mapping decision sample $M_k$ in a training iteration of the policy, we need to compress the target DNN to obtain the accuracy and latency performance for the calculation of the reward $R(\mathcal{M}_k)$. The latency is obtained via deploying pruned model with compiler code generation on target device and measuring the real execution time. To accelerate the policy training, we adopt magnitude-based, one-shot pruning and early stopping for faster accuracy evaluation during the policy training process. More specifically, once a mapping $\mathcal{M}_k$ is obtained, we conduct a one-shot pruning for each layer of the DNN based on the weight magnitude and retrain the DNN for two epochs to regain accuracy.
This partially regained accuracy can be used to predict the final model accuracy and compare the performance between different schemes~\cite{zhong2018practical, tan2019mnasnet}.
Furthermore, as compiler code generation and latency measurement do not depend on absolute weight values and are faster than DNN training, we overlap the compiler code generation and latency measurement with the accuracy evaluation of the pruned model.

\subsection{Rule-based Pruning Scheme Mapping Method} \label{sec:rule_based_method}

The advantage of the search-based method is that it can find the globally close-to-optimal pruning configurations for each of the layers in a given DNN. While it works perfectly for small DNN models, the searching overheads increases exponentially when the models size increases, making it unsuitable for large-size DNN models.
Therefore, we design a \textbf{\emph{training-free}} \emph{rule-based} method that maps the best-suited pruning schemes in a layer-wise fashion to avoid the time-consuming search process for the best mapping. 
We consider the search-based solution as the performance upper-bound, and we target to make the rule-based method perform as well as the search-based one, yet highly efficient and practical.

\subsubsection{\textbf{Latency Model}} \label{sec:latency_model}
To obtain the latency performance without the pruning and retraining of the given DNN, we build latency models for different types of layers, e.g., 1$\times$1 CONV, 3$\times$3 CONV, 5$\times$5 CONV, and depthwise-3$\times$3 CONV, on the target device, e.g., Samsung S10 smartphone. Each latency model contains latency results for different settings, including block size, number of filters, input and output feature map size, pruning scheme, and compression rate. The results are measured on the target device by running test models with each setting for 100 runs. Each test model has 10 cascaded layers with the same setting. 
Since building the latency model does not involve DNN training, it will not take a very long time.
The testing time for each run of each setting is in milliseconds (ms) level. For instance, our latency model including 512 different layer settings can be built in around 30 minutes.
Such a building time is negligible compared to the DNN training or the searching process, which usually counts in days.
The latency model only needs to be built once for a target device and is universal to different DNN models.

\subsubsection{\textbf{Block-Size Selection}} \label{sec:block_size_selection}
Block size has a significant impact on the accuracy and hardware performance for block-based/block-punched pruning. A larger block size is typically more hardware-friendly and easier to leverage the built-in hardware acceleration, yet it may cause more severe accuracy degradation due to the coarse granularity. On the contrary, a smaller block size typically leads to higher accuracy but also increases the latency. An appropriate setting of the block size can achieve high accuracy as unstructured pruning (essentially with block size 1$\times$1) and high hardware acceleration performance as structured pruning (essentially with block size of whole weight tensor/matrix) simultaneously.  

To determine the proper block size for each layer without the requirement of a time-consuming training process, we consider to decouple the two optimization targets, i.e., accuracy and hardware performance. To minimize the impact of pruning on hardware performance, our rule-based method will first derive the inference latency of each block size from the \textbf{offline generated} latency models and normalize the latency (i.e., divide by the MACs of that layer). We introduce a latency threshold $\beta$, indicating the acceptable latency degradation range of the proposed general pruning regularity compared with coarse-grained structured pruning.
The value of $\beta$ can be adjusted according to the design requirement, and it can either be the same for the entire model or different for each layer. 
For example, $\beta = 20\%$ means that the inference speed of block-based/block-punched pruning
can be at most 20\% slower than structured pruning under the same compression rate. After the hardware performance-driven design is satisfied, we only need to consider the influence of block size on accuracy. 
As a smaller block size can provide a finer granularity in pruning and the consequent higher accuracy, the smallest valid block size that satisfies the $\beta$-degradation requirement is selected as the desired block size. This process \underline{depends on our latency model, and is free of training}.

\subsubsection{\textbf{3$\times$3 CONV Layer: Pattern or Block}}
\label{sec:3x3_layer}
For 3$\times$3 CONV layers, both pattern-based pruning and block-punched pruning can be applied. To map the best-suited pruning scheme, the problem is to compare the accuracy and inference latency of block-punched pruning and pattern-based pruning.

\begin{figure} [t]
     \centering
     \includegraphics[width=0.7\columnwidth]{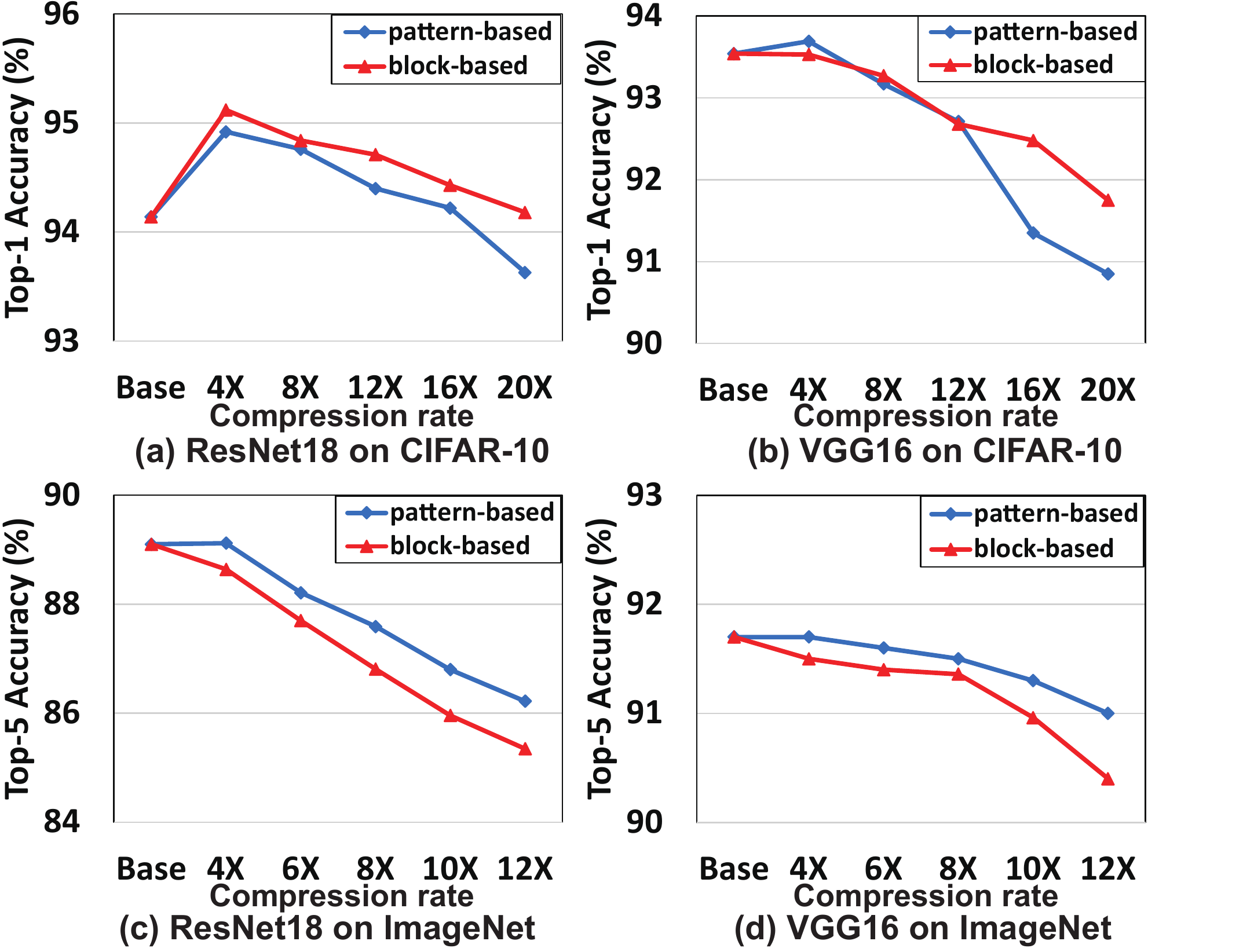}    
     \caption{The top-1/top-5 accuracy comparisons of pattern-based pruning and block-punched pruning (block size of 4$\times$16) under the same compression rates for ResNet-18 and VGG-16 on CIFAR-10 and ImageNet dataset.}
     \label{fig:res_pattern_vs_blk}
\end{figure}

\textbf{Accuracy Perspective:}
To investigate the accuracy of pattern-based pruning and block-punched pruning, we conduct comprehensive experiments on ResNet-18 and VGG-16 with CIFAR-10 and ImageNet dataset. 
Fig.~\ref{fig:res_pattern_vs_blk} (a) and (b) show an example of the comparison results on the CIFAR-10 dataset and the block size is set to 4$\times$16. 
Note that only 3$\times$3 CONV layers are pruned and non-3$\times$3 layers remain unpruned to provide a fair comparison. 
Here, the compression rate indicates the parameter reduction rate for each 3$\times$3 CONV layer. From the figure, we can make the following observations: 
i) block-punched pruning consistently shows comparable or higher accuracy for the pruned model under different compression rates on CIFAR-10 dataset; 
ii) both block-punched and pattern-based pruning achieve accuracy enhancement when the compression rate is relatively low, especially on ResNet-18. The reason is that pruning with a small compression rate can help mitigate the over-fitting problem.

The comparison results of pattern-based pruning and block-punched pruning on ImageNet dataset with different compression rates are shown in Fig.~\ref{fig:res_pattern_vs_blk} (c) and (d). 
Different from the observations on CIFAR-10 dataset, pattern-based pruning shows a better accuracy performance under various compression rate settings for both ResNet-18 and VGG-16. 

We attribute the different performance on the two datasets to:
(1) For tasks on easy datasets such as CIFAR-10 that can easily achieve higher than 90\% accuracy, the networks are generally overparameterized and both block-punched and pattern-based pruning schemes can achieve a high compression rate (e.g., >10$\times$) and significant acceleration without hurting the model generalization ability. Thus, the acceleration performance of the two pruning schemes becomes a more essential factor that contributes to the pruning scheme selection. 
Compared to pattern-based pruning, the block-based/block-punched pruning has a more strict constraint on the weight structure, benefiting hardware parallelism and hence a higher acceleration performance under the same compression rate. 
Therefore, block-based/block-punched pruning is more favorable for easier datasets.
(2) For tasks on harder datasets such as ImageNet that even the unpruned network can only achieve less than 80\% Top-1 accuracy, the pattern-based pruning scheme is more desirable than block-based/block-punched pruning on 3$\times$3 CONV layers. Because the patterns used by pattern-based pruning form the shape of Gaussian filter or Laplacian of Gaussian filter that can enhance the ability for feature extraction (as mentioned in Sec.~\ref{sec:weight_pruning}), which plays an important role in preserving accuracy under an accelerable compression rate.

Based on the above results, we make the following remark:
\begin{remark} \label{Remark:Remark1}
 For 3$\times$3 CONV layers, block-punched pruning is more suitable for tasks with easier datasets while pattern-based pruning suits tasks with harder datasets better.
\end{remark}
We will provide more discussions and verification of the remark in Section \ref{sec:framework_result}. 

\textbf{Latency Perspective:} Latency is the other important aspect in performance evaluation of a pruning scheme. 
From comprehensive comparative experiments conducted offline,
we have observed that under the same compression rate, 
the latency performance of block-punched pruning is better than pattern-punched pruning when the block size is large, but worse when the block size is small. 
The latency of these two pruning regularities mainly depends on which one can achieve a larger compression rate under the same accuracy.
Thus, latency is considered as a secondary criterion for the best-suited pruning scheme mapping in the rule-based method. More discussion will be provided in Section~\ref{sec:framework_result}.

\subsubsection{\textbf{3$\times$3 Depth-wise CONV Layer}}
\label{sec:DW3X3}

The 3$\times$3 depth-wise CONV layer (3$\times$3-DW) is widely used in current DNN designs such as the MobileNet family~\cite{sandler2018mobilenetv2}. 
It is a special case of 3$\times$3 CONV layer, which applies a 2-D depth filter at each depth level of the input tensor. 
Thus, both pattern-based pruning and block-punched pruning can be applied to 3$\times$3-DW layers theoretically. 
In our rule-based selection policy, we prefer not prune 3$\times$3-DW layers mainly for two reasons: (1) 3$\times$3-DW layers are computation- and memory-efficient; (2) 3$\times$3-DW layers are sensitive to pruning.

We use MobileNet-V2 on ImageNet as an example, 33\% of layers are 3$\times$3-DW layers, but they only contribute 6.9\% MACs and 1.7\% parameters in total. Pruning 3$\times$3-DW layers will not achieve a considerable gain even if all of them are pruned. 
On the other hand, the 3$\times$3-DW layers contribute 33\% of activations, making each weight in the 3$\times$3-DW layer more significant.
Moreover, in a regular 3$\times$3 CONV layer, one input (activation) channel will be filtered by multiple CONV kernels that come from different CONV filters and have different pruned locations, mitigating the damage of pruning on feature extraction.
On the contrary, in a 3$\times$3-DW layer, one input channel will only be filtered by one CONV kernel, which makes 3$\times$3-DW layers more sensitive to the pruning.

We conduct an ablation study about the impact of pruning 3$\times$3-DW on accuracy and overall pruning ratio. The results show that pruning 3$\times$3-DW layers will only slightly increase the pruning ratio while leading to a noticeable accuracy loss.
Our experiment results shown in Section~\ref{sec:block_related_results} indicate both pattern-based pruning and block-punched pruning lead to a non-negligible accuracy drop when applied to 3$\times$3-DW layers.
Therefore, our rule-based method does not map any pruning scheme to the 3$\times$3 depth-wise CONV layers.

\begin{figure} [t]
     \centering
     \includegraphics[width=0.5\columnwidth]{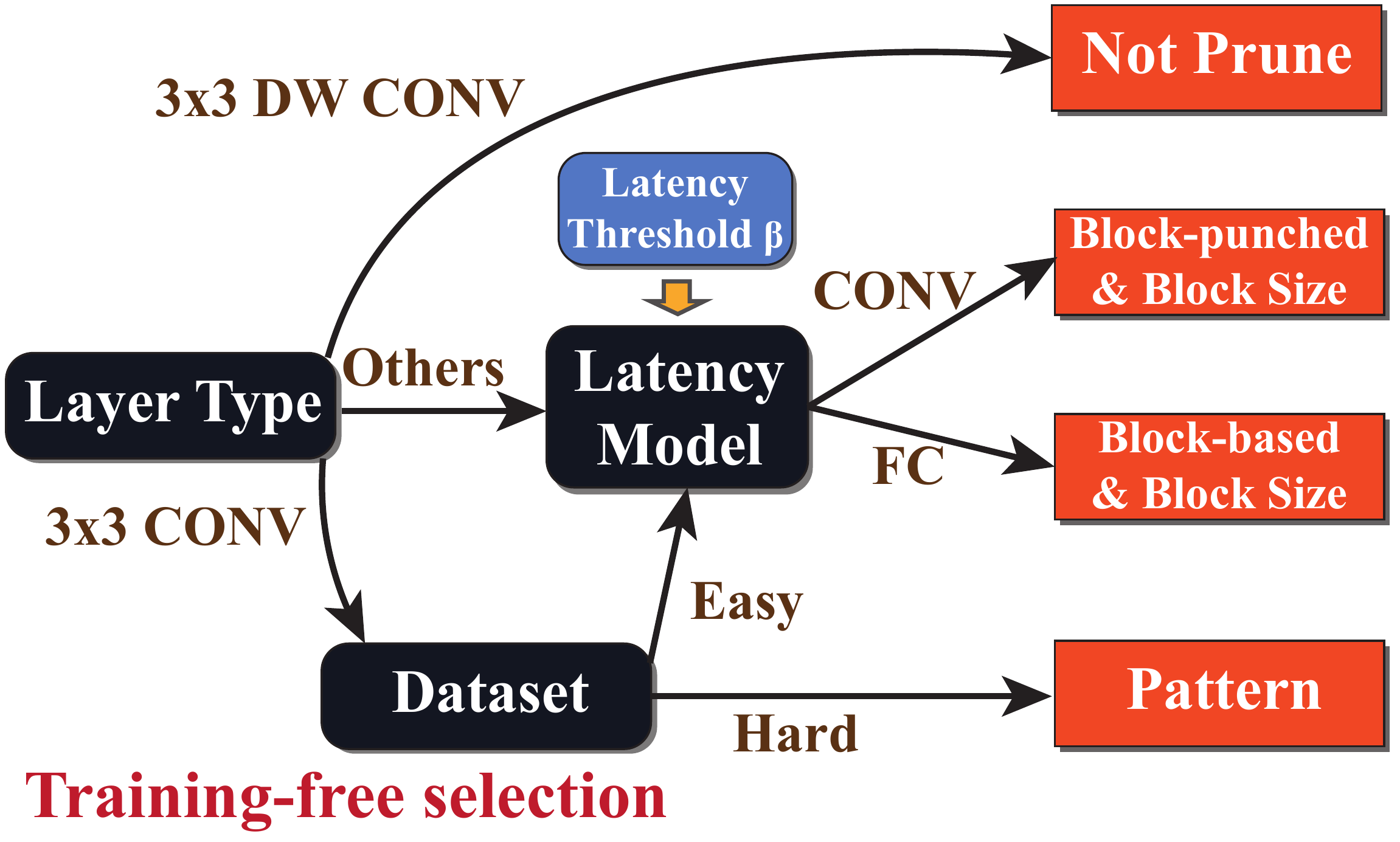}  
     \caption{Rule-based pruning scheme selection.}
     \label{fig:decision_tree}
\end{figure}

We summarize the workflow of the training-free rule-based method in Fig. \ref{fig:decision_tree}. For each layer of a given DNN, we first examine the layer type. If the layer is a 3$\times$3 depth-wise CONV layer, no pruning scheme is mapped. For 3$\times$3 CONV layers, the pruning regularity depends on the size of the dataset. Pattern-based pruning is mapped to 3$\times$3 CONV layers if the task has a large dataset, otherwise block-punched pruning is selected. The proposed general block-based/block-punched pruning is mapped to all other types of layers. When block-based/block-punched pruning is selected, the block size is determined according to an offline generated latency model with a latency threshold. We note that the entire mapping process, including the pruning regularity mapping and block size selection, \underline{is training-free without incurring any additional cost}.

\section{Evaluation}
\label{sec:evaluation}

\subsection{Methodology}
\textbf{Evaluation Objective:} (i) Show the effectiveness of the general, fine-grained structured pruning scheme and the corresponding compiler optimizations; 
(ii) compare the overall pruning scheme mapping framework with state-of-the-art DNN inference acceleration framework PatDNN \cite{niu2020patdnn} in terms of accuracy and latency. Note that PatDNN already outperforms other DNN inference frameworks including TVM \cite{chen2018tvm}, MNN \cite{Ali-MNN}, and Tensorflow-Lite \cite{TensorFlow-Lite}, thus the comparison with PatDNN is sufficient to show the effectiveness of our methods.

\underline{Our achieved speedup mainly comes from}: (i) our general, fine-grained structured pruning is applicable to all types of layers, which better compresses the model size and reduces the computation workload;
(ii) our compressed sparse matrix storage and associated compiler optimizations improve the computation regularity/parallelism, thus transforming the computation reduction to real performance gains;
(iii) our automatic pruning scheme mapping methods successfully map the best-suited pruning configurations to each layer, maximizing the compression rate while maintaining accuracy.

\noindent\textbf{DNN Models:} We evaluate on three mainstream DNNs, VGG-16 \cite{simonyan2014very}, ResNet-50 \cite{he2016deep}, and MobileNet-V2 \cite{sandler2018mobilenetv2}. They are trained on two representative datasets, CIFAR-10 and ImageNet \cite{deng2009imagenet}. We also conduct experiments on YOLOv4 \cite{bochkovskiy2020yolov4} with MS COCO dateset \cite{lin2014microsoft}. 

\noindent\textbf{Evaluation Platforms and Running Configurations:} All the evaluated models are trained on a server with 8 NVIDIA RTX 2080Ti GPUs. 
The training codes are implemented with the PyTorch API.
The latency is measured on the mobile GPU of an off-the-shelf Samsung Galaxy S10 smartphone, which has the Qualcomm Snapdragon 855 mobile platform with a Qualcomm Kryo 485 Octa-core CPU and a Qualcomm Adreno 640 GPU. Each test takes 50 runs on different inputs with 8 threads on CPU, and all pipelines on GPU. As different runs do not vary greatly, only the average time is reported for readability. All runs are tuned to the best configurations. 
We empirically choose the latency threshold $\beta=20\%$.

\begin{figure} [t!]
     \centering
     \includegraphics[width=0.7\columnwidth]{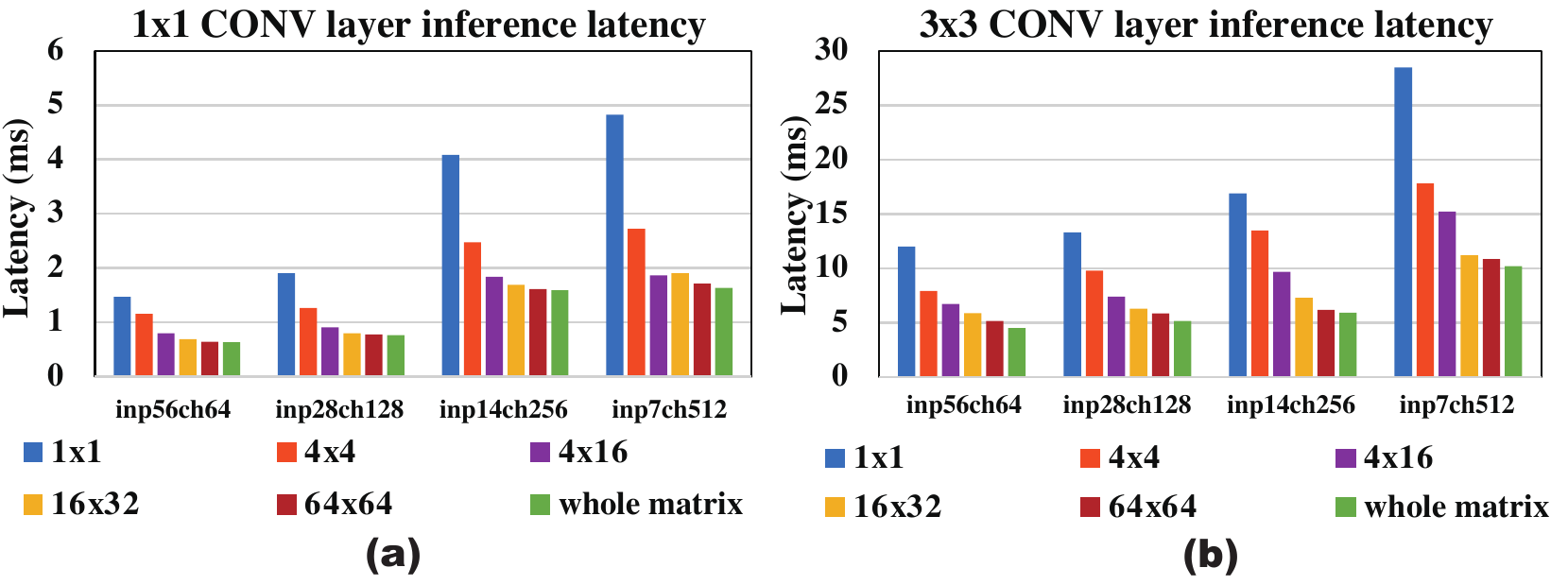}  
     \caption{{Latency of 1$\times$1 and 3$\times$3 CONV layer under different feature sizes and input/output channels.}}
     \label{fig:res_block_size}
\end{figure}

\subsection{Evaluations of Proposed Pruning Scheme}\label{sec:block_related_results}

We first evaluate the inference latency of block-punched pruning using different block sizes on 1$\times$1 and 3$\times$3 CONV layers with different layer sizes, as shown in Fig.~\ref{fig:res_block_size}.
The input feature map size of the testing CONV layers is set to 56$\times$56, 28$\times$28, 14$\times$14, and 7$\times$7 while the input/output channel size is set to 64, 128, 256, and 512. These configurations are commonly used in real DNN networks such as ResNet-50 and VGG-16 on ImageNet. And these configurations keep the MACs the same for all 1$\times$1/3$\times$3 CONV layers, which can help us observe the impact of different input feature map size and number of channels on latency better. 

From Fig.~\ref{fig:res_block_size} (a), we can see that the latency is reduced with a larger block size. However, the speedup gradually saturates. The reason is that the remaining weights in each block are more likely to be sufficient to exploit high hardware parallelism with larger block size. 
Another observation is that the layer inference latency increases for all block sizes as the size of the input feature map decreases and the number of input/output channels increases. The reason is that a smaller input feature map size lowers the reuse rate of each weight, causing hardware parallelism degradation. Similar observations can also be found in Figure~\ref{fig:res_block_size} (b), which shows the latency results for different 3$\times$3 CONV layers.


Similar results can also be observed on FC layers with block-based pruning. 
Fig.~\ref{fig:3x3_latency} (a) shows the latency comparisons on two FC layers. 
The size of the FC layer on the left-hand side is used as the first FC layer in VGG-16, while the right-hand side is the representative FC layer in BERT.
The latency of each FC layer is normalized to its own 1$\times$1 block size result. We can observe that for large FC layers, increasing the block size can reduce latency effectively, while the latency reduction achieved by increasing the block size gets saturated gradually in relatively small FC layers. 

\subsection{Automatic Pruning Scheme Mapping Methods Evaluations} \label{sec:framework_result} 

\begin{table}[!t]
\setlength\tabcolsep{2pt} 
\small
\centering
\caption{Comparison on YOLOv4.}
\label{table:yolov4}
\begin{tabular}{c|c|c|c|c}
\toprule
\hline
Pruning Scheme & \# Weights & Compres. Rate & mAP & FPS \\ \hline
Not Prune & 64.36M & 1$\times$ & 57.3  & 3.5  \\ \hline
Structured & 8.82M & 7.3$\times$ & 39.4 & 11.8 \\ \hline
Unstructured & 5.75M & 11.2$\times$ & 52.5& 7.6 \\ \hline
Pattern & 10.22M & *6.3$\times$ & 52.8 & 9.7 \\ \hline
Block & 10.38M & *6.2$\times$ & 52.4 & 9.1 \\ \hline
Block & 7.94M & 8.1$\times$ & 51.3 & 11.5 \\ \hline
Hybrid & 7.57M & 8.5$\times$ & 51.7 & 12.3 \\ \hline\bottomrule
\multicolumn{5}{l}{* Overall compression rate, but only 3$\times$3 CONV layers are pruned. } 
\end{tabular}
\vspace{-12pt}
\end{table}



\begin{table*}[t]
	\centering
		\begin{minipage}[c]{0.651\textwidth}
     \centering
     \includegraphics[width=0.89\columnwidth]{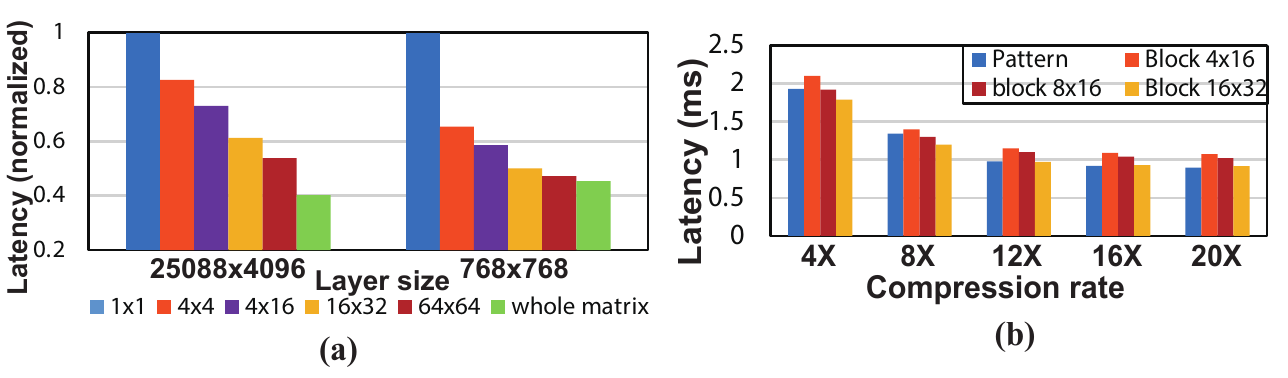}    
     \captionof{figure}{(a) Latency of two example FC layers. (b) Example of latency comparisons of 3$\times$3 CONV layer using pattern-based pruning and block-punched pruning.}
     \label{fig:3x3_latency}
	\end{minipage}\hspace{5pt}
\begin{minipage}[c]{0.33\textwidth}
\setlength\tabcolsep{2pt}
\scriptsize
\centering
\caption{Accuracy comparison ($\Delta$ acc.) of applying pattern-based pruning and block-punched pruning to the depth-wise 3$\times$3 CONV layers in MobileNetV2.}
\label{table:dw_conv}
\begin{tabular}{c|c|c}
\toprule\hline
 & CIFAR-10 & CIFAR-100 \\ \hline
Compres. rate & 7.19$\times$-\textgreater{}8.12$\times$ & 2.78$\times$-\textgreater{}2.91$\times$ \\ \hline
Pattern-based  & -0.4 & -0.9 \\ \hline
Block-based  & -1.01 & -1.51 \\ \hline\bottomrule

\end{tabular}
\end{minipage}
\end{table*}

\begin{table}[t]
\centering
\scriptsize
\caption{Comparison with PatDNN on CIFAR-10 dataset (top-1 acc.) and ImageNet dataset (top-1/top-5 acc.).}
\label{table:prune_results}
\begin{tabular}{c|c|c|c|c|c|c|c|c}
\toprule
\hline
Network & Method & \begin{tabular}[c]{@{}c@{}}Pruning\\ scheme\end{tabular} & Pruned layers & \begin{tabular}[c]{@{}c@{}}Original\\ acc. (\%)\end{tabular} & \begin{tabular}[c]{@{}c@{}}Compres.\\ rate\end{tabular} & \begin{tabular}[c]{@{}c@{}}Acc.\\ drop (\%)\end{tabular} & \begin{tabular}[c]{@{}c@{}}Latency\\ (ms)\end{tabular} &
\begin{tabular}[c]{@{}c@{}}MACs\\ \end{tabular} \\ \hline
\multicolumn{8}{c}{CIFAR-10} \\ \hline
\multirow{3}{*}{ResNet-50} & PatDNN & Pattern & 3x3 CONV & 95.6 & 1.57$\times$ & -1.0 & 10.44 &1.9G\\ \cline{2-9} 
&  {Rule-based} &  {Block} &  {3x3 CONV, 1x1 CONV} &  {95.6} &  {11.51$\times$} &  {0.1} &  {4.25} &0.6G\\ \cline{2-9} 
& Search-based & Hybrid & 3x3 CONV, 1x1 CONV & 95.6 & 11.88$\times$ & 0.1 &  4.20&0.6G\\  \hline
 
\multirow{3}{*}{VGG-16} & PatDNN & Pattern & 3x3 CONV & 93.9 & 8.0$\times$ & -0.4 & 2.59 & 73M \\ \cline{2-9} 
 &  {Rule-based} &  {Block} &  {3x3 CONV} &  {93.9} &  {12.38$\times$} &  {-0.3} &  {2.02} & 59M \\ \cline{2-9} 
 & Search-based & Hybrid & 3x3 CONV & 93.9 & 12.50$\times$ & -0.3 &  2.00 & 58M\\  \hline
 
\multirow{3}{*}{MobileNetV2} & PatDNN & Pattern & 3x3 DW CONV & 94.6 & 1.01$\times$ & -0.1 & 3.63 & 296M\\ \cline{2-9} 
 &  {Rule-based} &  {Block} &  {1x1 CONV} &  {94.6} &  {7.53$\times$} &  {0.2} &  {1.86}& 89M\\ \cline{2-9} 
 & Search-based& Block & 1x1 CONV & 94.6 & 7.54$\times$ & 0.1 &  1.86& 89M\\  \hline
 
\multicolumn{8}{c}{ImageNet} \\ \hline
\multirow{3}{*}{ResNet-50} & PatDNN & Pattern & 3x3 CONV & 76.1/92.8 & 1.56$\times$ & ---/-0.2 & 29.89 & 3.0G\\ \cline{2-9} 
 &  {Rule-based} &  {Hybrid} &  {3x3 CONV, 1x1 CONV} &  {76.1/92.8} &  {4.37$\times$} &  {0.3/0.1} &  {17.26} & 1.6G\\ \cline{2-9} 
  & Search-based& Hybrid & 3x3 CONV, 1x1 CONV & 76.1/92.8 & 4.41$\times$ & 0.1/0 &  17.22 & 1.6G\\  \hline
  
\multirow{3}{*}{VGG-16} & PatDNN & Pattern & 3x3 CONV & 74.5/91.7 & 8.0$\times$ & ---/0.1 & 18.91 & 3.8G \\ \cline{2-9} 
 &  {Rule-based} &  {Pattern} &  {3x3 CONV} &  {74.5/91.7} &  {8.22$\times$} &  {0.2/0.1} &  {18.17} & 3.5G \\ \cline{2-9} 
  & Search-based& Pattern & 3x3 CONV & 74.5/91.7 & 8.22$\times$ & 0.2/0.1 &  18.17& 3.5G\\  \hline
  
\multirow{3}{*}{MobileNetV2} & PatDNN & Pattern & 3x3 DW CONV & 71.0/90.3 & 1.01$\times$ & ---/0 & 4.90 & 300M\\ \cline{2-9} 
 &  {Rule-based} &  {Block} &  {1x1 CONV} &  {71.0/90.3} &  {1.76$\times$} &  {0.5/0.3} &  {3.98} & 177M \\ \cline{2-9} 
  & Search-based & Block & 1x1 CONV & 71.0/90.3 & 1.82$\times$ & 0.5/0.3 & 3.90 & 165M \\  \hline
 \bottomrule
\end{tabular}
\end{table}

\subsubsection{Accuracy Analysis on Pattern-based Pruning and Block-Punched Pruning} \label{sec:yolo}
From the results on ResNet-18 and VGG-16 with CIFAR-10 and ImageNet dataset, we make Remark \ref{Remark:Remark1}. We further examine the remark on YOLOv4 with MS COCO dataset, which can be reasonably regarded as difficult task, as shown in Table \ref{table:yolov4}. The compression rate refers to the compression rate of the entire model and the block size is 4$\times$16. 
When only 3$\times$3 CONV layers are pruned, pattern-based pruning achieves a higher mean average precision (mAP), which matches the remark that pattern-based pruning suits tasks with larger dataset better on 3$\times$3 CONV layers. 
However, current pattern-based pruning is only applicable to 3$\times$3 layers, limiting the compression performance. With the proposed general pruning scheme applicable to different layers, we achieve an 8.1$\times$ compression rate with 51.3 mAP and 11.5 frames per second (FPS). A hybrid pruning scheme by mapping pattern-based pruning to 3$\times$3 CONV layers and block-based/block-punched pruning to all the other layers can further achieve an 8.5$\times$ compression rate with 51.7 mAP and 12.3 FPS. 
We also show the results of unstructured pruning and structured pruning, which achieve 52.5 mAP and 39.4 mAP, and 7.6 FPS and 11.8 FPS, respectively. It can be observed that our hybrid scheme method is 1.62$\times$ faster than unstructured pruning while maintaining comparable accuracy. When compared to structured pruning, our hybrid scheme method achieve much higher accuracy and is also slightly faster than structured pruning at the same time.
This further strengthens the advantage of our proposed method.

\subsubsection{Latency Analysis on Pattern-based Pruning and Block-Punched Pruning}
We conduct comprehensive comparative experiments offline to analyze the latency performance of pattern-based pruning and block-punched pruning to determine the best-suited pruning scheme for 3$\times$3 CONV layers.
Fig.~\ref{fig:3x3_latency} (b) shows an example of the latency comparisons for a 3$\times$3 CONV layer with 28$\times$28 input feature map size and 128 input/output channels under different compression rates. 
Under 4$\times$ and 8$\times$ compression, pattern-based pruning has similar latency performance to block-punched pruning with a block size of 8$\times$16.
When the compression rate is higher than 12$\times$, pattern-based pruning has similar speed as block-punched pruning with a block size of 16$\times$32. However, the latency difference between pattern-based pruning and block-punched pruning is minor, as we discussed in Section~\ref{sec:3x3_layer}, thus we consider latency performance as a secondary criterion in rule-based pruning scheme mapping method.

\subsubsection{Ablation Study on 3$\times$3 Depth-wise CONV Layer}
As mentioned in Section~\ref{sec:DW3X3}, 3$\times$3 depth-wise CONV layers usually only account for a small portion of weights and computations, and they play an important role in capturing spatial correlations in DNNs~\cite{Chollet_2017}, thus we propose not to prune 3$\times$3 depth-wise CONV layers.
Table~\ref{table:dw_conv} shows the accuracy results of applying pattern-based pruning and block-punched pruning to 3$\times$3 depth-wise CONV layers in MobileNetV2. 
Here we use the baseline models that all the 1$\times$1 CONV layers are pruned by block-punched pruning with compression rate 7.19$\times$ and 2.78$\times$ for CIFAR-10 and CIFAR-100, respectively. 
Then, on top of the pruned model, we apply an extra 2.22$\times$ pattern-based/block-punched pruning only for the 3$\times$3 depth-wise CONV layers and compare the final accuracy.
The results show that the overall compression rate only increases slightly, but there is non-negligible accuracy drop for pattern-based pruning and block-based pruning.
Thus, our rule-based pruning scheme mapping method will not map any pruning scheme for 3$\times$3 depth-wise CONV layers.

\subsubsection{Evaluations of Automatic Pruning Scheme Mapping Methods}
We compare the search-based and rule-based methods with the state-of-the-art end-to-end inference framework PatDNN \cite{niu2020patdnn}, which uses pattern-based pruning with ADMM pruning algorithm.
The comparison results are shown in Table \ref{table:prune_results}. Here, the compression rate refers to the parameter reduction rate of the CONV layers. The accuracy for ImageNet dataset indicates the top-5 accuracy. 

The configurations of the search-based method are obtained using 5 GPU servers, and take 3 and 9 days for CIFAR-10 and ImageNet models, respectively, which is acceptable for RL-based search methods \cite{zoph2016neural, tan2019mnasnet}.
We use search-based method to provide a close-to-optimal result, which indicates the performance upper-bound.
Accelerating the search process is not the main concern of our work, and our search process can be accelerated by adopting fast evaluation techniques such as 
Bayesian Optimization~\cite{klein2017fast,chen2018bayesian}.

For ResNet-50 on CIFAR-10, rule-based method can achieve an 11.51$\times$ compression rate with only 0.1\% accuracy drop, significantly higher than the results obtained by PatDNN. The reason for the limited performance of PatDNN is that only 44.3\% of the parameters of ResNet-50 are in the 3$\times$3 CONV layers that can be pruned with pattern-based pruning, as shown in Fig.~\ref{fig:para_ratio}. Our rule-based method, on the other hand, maps the flexible block-punched pruning that can be applied to CONV layers with different kernel sizes, thus achieving a much higher compression rate. Search-based method reaches a slightly higher compression rate and minor latency reduction compared with the rule-based method. 

With the automatic mapping of block-punched pruning and block size provided by the rule-based method and compression rate derived by the reweighted pruning algorithm, we reach a 12.38$\times$ compression rate with 0.3\% accuracy improvement on VGG-16 for CIFAR-10 dataset. Still, search-based method renders a slightly better performance than rule-based method.

For MobileNet-V2, mapping block-based pruning with optimized block size on 1$\times$1 CONV layers by rule-based method achieves a 7.53$\times$ compression rate with only 0.2\% accuracy drop. The compression rate is much higher than PatDNN, as pattern-based pruning cannot be applied to 1$\times$1 CONV, and 3$\times$3 depth-wise CONV layers only account for 1.9\% of the parameters in the model. The performance difference between rule-based method and search-based method is negligible.
 
Different from CIFAR-10, pattern-based pruning has a better accuracy performance on tasks with large datasets like ImageNet as discussed in Remark \ref{Remark:Remark1}. Hence, rule-based method maps pattern-based pruning to 3$\times$3 CONV layers and block-punched pruning with optimized block size to the remaining layers. For ResNet-50 on ImageNet, rule-based method can reach a 4.37$\times$ compression rate with only 0.1\% accuracy loss, and 1.73$\times$ speedup on mobile GPU over PatDNN. 

For VGG-16 on ImageNet, both rule-based method and search-based method maps pattern-based pruning to all the 3$\times$3 layers with the reweighted dynamic regularization algorithm, and achieves a 8.22$\times$ compression rate with only 0.1\% accuracy loss, which outperforms PatDNN. As all methods adopt pattern-based pruning, the performance difference between our methods and PatDNN is attributed to the pruning algorithm. With the reweighted pruning aglorithm, our method has the advantage of determining the compression rate for each layer automatically while PatDNN is based on ADMM and requires the manual setting of the compression rate for each layer. For MobileNet-V2 on ImageNet dataset, both rule-based method and search-based method map block-punched pruning to 1$\times$1 CONV layers, and reaches a 1.76$\times$ compression rate and 1.82$\times$ compression rate, respectively.

We also compare our method with other representative model compression techniques including NetAdapt ~\cite{Yang_2018}, ChamNet~\cite{Dai_2019}, AMC ~\cite{he2018amc}, AutoSlim \cite{liu2019autoslim}, and MetaPruning ~\cite{Liu_2019} on the ImageNet dataset, and the results are shown in Table \ref{table:compression_technique_compare}. At 200M MACs level, our rule-based method achieves the same accuracy as AMC with fewer MACs. Our method also outperforms the 0.75$\times$ channel scaled MoileNetV2 in both accuracy and MACs. At 150M MACs level, the model obtained by our rule-based model achieves the highest top-1 accuracy with similar MACs compared with AutoSlim and the 0.5$\times$ channel scaled MobileNetV1. 

Combining all the results, we can see that both the rule-based and the search-based method significantly outperform PatDNN. Rule-based method can provide pruned models with similar accuracy and latency performance as search-based method, and avoids the policy training process, thus is more useful in practice. Moreover, with the assist of our compiler optimization, both methods can easily achieve real-time DNN inference (less than 33ms) on all models mentioned above.

\begin{table}[]
\centering
\small
\caption{The comparisons with models obtained by various model compression techniques on ImageNet. } \label{table:compression_technique_compare}
\begin{tabular}{clcc}
\toprule\hline
Group                      & Model                    & MACs   & Top-1 Acc \\ \hline
\multirow{3}{*}{300M MACs}  & MobileNetV2 1.0$\times$  & 300M   & 71.0\%    \\ \cline{2-4} 
                           & NetAdapt-MobileNetV1~\cite{Yang_2018}     & 284.3M & 69.1\%    \\ \cline{2-4} 
                           & ChamNet-B~\cite{Dai_2019}                & 323M   & 73.8\%    \\ \hline
\multirow{5}{*}{200M MACs} & MobileNetV2 0.75$\times$ & 209M   & 69.8\%    \\ \cline{2-4} 
                           & AMC-MobileNetV2~\cite{he2018amc}          & 211M   & 70.8\%    \\ \cline{2-4} 
                           & AutoSlim-MobileNetV2~\cite{liu2019autoslim}     & 207M   & 73\%      \\ \cline{2-4} 
                           & MetaPruning-MobileNetV2~\cite{Liu_2019}  & 217M   & 71.2\%    \\ \cline{2-4} 
                           & Ours (Rule-based)                    & 203M       & 70.8\%    \\ \hline
\multirow{4}{*}{150M MACs} & MobileNetV1 0.5$\times$  & 150M   & 63.3\%    \\ \cline{2-4} 
                           & AutoSlim-MobileNetV1~\cite{liu2019autoslim}    & 150M   & 67.9\%    \\ \cline{2-4} 
                           & Ours (Rule-based)                    & 177M   & 70.5\%   \\\cline{2-4}
                           & Ours  (Rule-based)                   & 151M   & 69.8\%   \\
                           \hline\bottomrule
\end{tabular}
\end{table}

\subsubsection{Portability Evaluation on Different Platforms}
We further evaluate the portability of our proposed rule-based pruning scheme mapping method on different mobile platforms. Three tested platforms are Samsung Galaxy S10, S20, and S21, respectively. They are equipped with different types of chipsets and mobile GPUs. The detailed hardware specifications are shown in Table~\ref{table:hardware_spec}.
Table~\ref{table:portability_test} shows the portability evaluation results on the three platforms using our rule-based pruning scheme mapping method. We use VGG16 network and test on CIFAR10 and ImageNet dataset, respectively. 
We build latency model for each platform and use the same latency threshold $\beta=20\%$. 
It can be observed that our rule-based method can consistently achieve high model accuracy and leverages the better hardware for a higher inference speed, which illustrates the stability of our reweighted pruning algorithm and the effectiveness and portability of our rule-based method.

\begin{table}[]
\centering
\small
\caption{Hardware specifications of platforms for portability evaluation.} \label{table:hardware_spec}
\begin{tabular}{cccc}
\toprule
\hline
Model & Chipset & GPU & RAM \\ \hline
Samsung Galaxy S10 & Qualcomm Snapdragon 855  & Adreno 640 & 8GB \\ \hline
Samsung Galaxy S20 & Qualcomm Snapdragon 865 & Adreno 650 & 12GB \\ \hline
Samsung Galaxy S21 & Qualcomm Snapdragon 888 & Adreno 666 & 8GB \\ \hline
\bottomrule
\end{tabular}
\end{table}

\begin{table}[]
\small
\caption{Portability evaluation on different platforms using rule-based method on VGG16.}
\label{table:portability_test}
\begin{tabular}{cccccc}
\toprule\hline
Dataset & Platform & \begin{tabular}[c]{@{}c@{}}Compres.\\ rate\end{tabular} & MACs & \begin{tabular}[c]{@{}c@{}}Top-1\\ acc.\end{tabular} & \begin{tabular}[c]{@{}c@{}}Latency\\ (ms)\end{tabular} \\ \hline
\multirow{3}{*}{CIFAR10} & Galaxy S10 & 12.38$\times$ & 59M & 94.2\% & 2.02 \\ \cline{2-6} 
 & Galaxy S20 & 12.06$\times$ & 62M & 94.1\% & 1.85 \\ \cline{2-6} 
 & Galaxy S21 & 12.12$\times$ & 61M & 94.2\% & 1.65 \\ \hline
\multirow{3}{*}{ImageNet} & Galaxy S10 & 8.22$\times$ & 3.5G & 74.3\% & 18.17 \\ \cline{2-6} 
 & Galaxy S20 & 8.12$\times$ & 3.4G & 74.5\% & 16.23 \\ \cline{2-6} 
 & Galaxy S21 & 8.15$\times$ & 3.4G & 74.5\% & 15.12 \\ \hline\bottomrule
\end{tabular}
\end{table}

\section{Conclusion}
We propose a general pruning scheme with fine-grained structured pruning regularity and reweighted dynamic pruning algorithm. 
Compiler optimizations are introduced to extract the structure information and exploit hardware parallelism. We further probe into the new problem of mapping the best-suited pruning scheme for each layer of any given DNN and propose two automatic pruning scheme mapping methods. Experimental results demonstrate the effectiveness of the proposed pruning scheme and pruning scheme mapping methods.

\begin{acks}  

This research is partially funded by National   Science Foundation CCF-1919117, CNS-1909172, CCF-2047516 (CAREER), and
CCF-1901378, and Jeffress Trust Awards in Interdisciplinary Research to William \& Mary.
Any opinions, findings, and conclusions or recommendations expressed in this material are those of the authors and do not necessarily reflect the views of NSF or Thomas F. and Kate Miller Jeffress Memorial Trust.
\end{acks}

\bibliographystyle{ACM-Reference-Format}
\bibliography{references}


\begin{thebibliography}{88}


\ifx \showCODEN    \undefined \def \showCODEN     #1{\unskip}     \fi
\ifx \showDOI      \undefined \def \showDOI       #1{#1}\fi
\ifx \showISBNx    \undefined \def \showISBNx     #1{\unskip}     \fi
\ifx \showISBNxiii \undefined \def \showISBNxiii  #1{\unskip}     \fi
\ifx \showISSN     \undefined \def \showISSN      #1{\unskip}     \fi
\ifx \showLCCN     \undefined \def \showLCCN      #1{\unskip}     \fi
\ifx \shownote     \undefined \def \shownote      #1{#1}          \fi
\ifx \showarticletitle \undefined \def \showarticletitle #1{#1}   \fi
\ifx \showURL      \undefined \def \showURL       {\relax}        \fi
\providecommand\bibfield[2]{#2}
\providecommand\bibinfo[2]{#2}
\providecommand\natexlab[1]{#1}
\providecommand\showeprint[2][]{arXiv:#2}

\bibitem[\protect\citeauthoryear{??}{Ten}{[n.d.]}]%
        {TensorFlow-Lite}
 \bibinfo{year}{[n.d.]}\natexlab{}.
\newblock \bibinfo{howpublished}{https://www.tensorflow.org/mobile/tflite/}.
\newblock


\bibitem[\protect\citeauthoryear{??}{Ali}{[n.d.]}]%
        {Ali-MNN}
 \bibinfo{year}{[n.d.]}\natexlab{}.
\newblock \bibinfo{howpublished}{\url{https://github.com/alibaba/MNN}}.
\newblock


\bibitem[\protect\citeauthoryear{??}{Pyt}{[n.d.]}]%
        {Pytorch-Mobile}
 \bibinfo{year}{[n.d.]}\natexlab{}.
\newblock \bibinfo{howpublished}{https://pytorch.org/mobile/home}.
\newblock


\bibitem[\protect\citeauthoryear{Ashari, Tatikonda, Boehm, Reinwald, Campbell,
  Keenleyside, and Sadayappan}{Ashari et~al\mbox{.}}{2015}]%
        {ashari2015optimizing}
\bibfield{author}{\bibinfo{person}{Arash Ashari}, \bibinfo{person}{Shirish
  Tatikonda}, \bibinfo{person}{Matthias Boehm}, \bibinfo{person}{Berthold
  Reinwald}, \bibinfo{person}{Keith Campbell}, \bibinfo{person}{John
  Keenleyside}, {and} \bibinfo{person}{P Sadayappan}.}
  \bibinfo{year}{2015}\natexlab{}.
\newblock \showarticletitle{On optimizing machine learning workloads via kernel
  fusion}.
\newblock \bibinfo{journal}{\emph{ACM SIGPLAN Notices}} \bibinfo{volume}{50},
  \bibinfo{number}{8} (\bibinfo{year}{2015}), \bibinfo{pages}{173--182}.
\newblock


\bibitem[\protect\citeauthoryear{Bezanson, Edelman, Karpinski, and
  Shah}{Bezanson et~al\mbox{.}}{2017}]%
        {bezanson2017julia}
\bibfield{author}{\bibinfo{person}{Jeff Bezanson}, \bibinfo{person}{Alan
  Edelman}, \bibinfo{person}{Stefan Karpinski}, {and} \bibinfo{person}{Viral~B
  Shah}.} \bibinfo{year}{2017}\natexlab{}.
\newblock \showarticletitle{Julia: A fresh approach to numerical computing}.
\newblock \bibinfo{journal}{\emph{SIAM review}} \bibinfo{volume}{59},
  \bibinfo{number}{1} (\bibinfo{year}{2017}), \bibinfo{pages}{65--98}.
\newblock


\bibitem[\protect\citeauthoryear{Bochkovskiy, Wang, and Liao}{Bochkovskiy
  et~al\mbox{.}}{2020}]%
        {bochkovskiy2020yolov4}
\bibfield{author}{\bibinfo{person}{Alexey Bochkovskiy},
  \bibinfo{person}{Chien-Yao Wang}, {and} \bibinfo{person}{Hong-Yuan~Mark
  Liao}.} \bibinfo{year}{2020}\natexlab{}.
\newblock \showarticletitle{Yolov4: Optimal speed and accuracy of object
  detection}.
\newblock \bibinfo{journal}{\emph{arXiv preprint arXiv:2004.10934}}
  (\bibinfo{year}{2020}).
\newblock


\bibitem[\protect\citeauthoryear{Boehm, Reinwald, Hutchison, Evfimievski, and
  Sen}{Boehm et~al\mbox{.}}{2018}]%
        {boehm2018optimizing}
\bibfield{author}{\bibinfo{person}{Matthias Boehm}, \bibinfo{person}{Berthold
  Reinwald}, \bibinfo{person}{Dylan Hutchison}, \bibinfo{person}{Alexandre~V
  Evfimievski}, {and} \bibinfo{person}{Prithviraj Sen}.}
  \bibinfo{year}{2018}\natexlab{}.
\newblock \showarticletitle{On optimizing operator fusion plans for large-scale
  machine learning in systemml}.
\newblock \bibinfo{journal}{\emph{arXiv preprint arXiv:1801.00829}}
  (\bibinfo{year}{2018}).
\newblock


\bibitem[\protect\citeauthoryear{Cai, Zhu, and Han}{Cai et~al\mbox{.}}{2018}]%
        {cai2018proxylessnas}
\bibfield{author}{\bibinfo{person}{Han Cai}, \bibinfo{person}{Ligeng Zhu},
  {and} \bibinfo{person}{Song Han}.} \bibinfo{year}{2018}\natexlab{}.
\newblock \showarticletitle{Proxylessnas: Direct neural architecture search on
  target task and hardware}.
\newblock \bibinfo{journal}{\emph{arXiv preprint arXiv:1812.00332}}
  (\bibinfo{year}{2018}).
\newblock


\bibitem[\protect\citeauthoryear{Cai, Li, Yuan, Niu, Li, Tang, Ren, and
  Wang}{Cai et~al\mbox{.}}{2021}]%
        {cai2021yolobile}
\bibfield{author}{\bibinfo{person}{Yuxuan Cai}, \bibinfo{person}{Hongjia Li},
  \bibinfo{person}{Geng Yuan}, \bibinfo{person}{Wei Niu},
  \bibinfo{person}{Yanyu Li}, \bibinfo{person}{Xulong Tang},
  \bibinfo{person}{Bin Ren}, {and} \bibinfo{person}{Yanzhi Wang}.}
  \bibinfo{year}{2021}\natexlab{}.
\newblock \showarticletitle{YOLObile: Real-Time Object Detection on Mobile
  Devices via Compression-Compilation Co-Design}. In
  \bibinfo{booktitle}{\emph{Proceedings of the AAAI Conference on Artificial
  Intelligence}}, Vol.~\bibinfo{volume}{35}. \bibinfo{pages}{955--963}.
\newblock


\bibitem[\protect\citeauthoryear{Candes, Wakin, and Boyd}{Candes
  et~al\mbox{.}}{2008}]%
        {candes2008enhancing}
\bibfield{author}{\bibinfo{person}{Emmanuel~J Candes},
  \bibinfo{person}{Michael~B Wakin}, {and} \bibinfo{person}{Stephen~P Boyd}.}
  \bibinfo{year}{2008}\natexlab{}.
\newblock \showarticletitle{Enhancing sparsity by reweighted $\ell_{1}$
  minimization}.
\newblock \bibinfo{journal}{\emph{Journal of Fourier analysis and
  applications}} \bibinfo{volume}{14}, \bibinfo{number}{5-6}
  (\bibinfo{year}{2008}), \bibinfo{pages}{877--905}.
\newblock


\bibitem[\protect\citeauthoryear{Chen, Moreau, Jiang, Zheng, Yan, Shen, Cowan,
  Wang, Hu, Ceze, Guestrin, and Krishnamurthy}{Chen et~al\mbox{.}}{2018b}]%
        {chen2018tvm}
\bibfield{author}{\bibinfo{person}{Tianqi Chen}, \bibinfo{person}{Thierry
  Moreau}, \bibinfo{person}{Ziheng Jiang}, \bibinfo{person}{Lianmin Zheng},
  \bibinfo{person}{Eddie Yan}, \bibinfo{person}{Haichen Shen},
  \bibinfo{person}{Meghan Cowan}, \bibinfo{person}{Leyuan Wang},
  \bibinfo{person}{Yuwei Hu}, \bibinfo{person}{Luis Ceze},
  \bibinfo{person}{Carlos Guestrin}, {and} \bibinfo{person}{Arvind
  Krishnamurthy}.} \bibinfo{year}{2018}\natexlab{b}.
\newblock \showarticletitle{TVM: An automated end-to-end optimizing compiler
  for deep learning}. In \bibinfo{booktitle}{\emph{the USENIX Symposium on
  Operating Systems Design and Implementation (OSDI)}}.
\newblock


\bibitem[\protect\citeauthoryear{Chen, Huang, Wang, Antonoglou, Schrittwieser,
  Silver, and de~Freitas}{Chen et~al\mbox{.}}{2018a}]%
        {chen2018bayesian}
\bibfield{author}{\bibinfo{person}{Yutian Chen}, \bibinfo{person}{Aja Huang},
  \bibinfo{person}{Ziyu Wang}, \bibinfo{person}{Ioannis Antonoglou},
  \bibinfo{person}{Julian Schrittwieser}, \bibinfo{person}{David Silver}, {and}
  \bibinfo{person}{Nando de Freitas}.} \bibinfo{year}{2018}\natexlab{a}.
\newblock \showarticletitle{Bayesian optimization in alphago}.
\newblock \bibinfo{journal}{\emph{arXiv preprint arXiv:1812.06855}}
  (\bibinfo{year}{2018}).
\newblock


\bibitem[\protect\citeauthoryear{Chollet}{Chollet}{2017}]%
        {Chollet_2017}
\bibfield{author}{\bibinfo{person}{Fran{\c{c}}ois Chollet}.}
  \bibinfo{year}{2017}\natexlab{}.
\newblock \showarticletitle{Xception: Deep learning with depthwise separable
  convolutions}. In \bibinfo{booktitle}{\emph{Proceedings of the IEEE
  conference on computer vision and pattern recognition (CVPR)}}.
\newblock


\bibitem[\protect\citeauthoryear{Dai, Jia, Vajda, Uyttendaele, Jha, Zhang, Wu,
  Yin, Sun, Wang, and et~al.}{Dai et~al\mbox{.}}{2019a}]%
        {Dai_2019}
\bibfield{author}{\bibinfo{person}{Xiaoliang Dai}, \bibinfo{person}{Yangqing
  Jia}, \bibinfo{person}{Peter Vajda}, \bibinfo{person}{Matt Uyttendaele},
  \bibinfo{person}{Niraj~K. Jha}, \bibinfo{person}{Peizhao Zhang},
  \bibinfo{person}{Bichen Wu}, \bibinfo{person}{Hongxu Yin},
  \bibinfo{person}{Fei Sun}, \bibinfo{person}{Yanghan Wang}, {and}
  \bibinfo{person}{et al.}} \bibinfo{year}{2019}\natexlab{a}.
\newblock \showarticletitle{ChamNet: Towards Efficient Network Design Through
  Platform-Aware Model Adaptation}.
\newblock \bibinfo{journal}{\emph{2019 IEEE/CVF Conference on Computer Vision
  and Pattern Recognition (CVPR)}} (\bibinfo{date}{Jun} \bibinfo{year}{2019}).
\newblock
\showISBNx{9781728132938}
\urldef\tempurl%
\url{https://doi.org/10.1109/cvpr.2019.01166}
\showDOI{\tempurl}


\bibitem[\protect\citeauthoryear{Dai, Yin, and Jha}{Dai et~al\mbox{.}}{2019b}]%
        {dai2019grow}
\bibfield{author}{\bibinfo{person}{Xiaoliang Dai}, \bibinfo{person}{Hongxu
  Yin}, {and} \bibinfo{person}{Niraj~K Jha}.} \bibinfo{year}{2019}\natexlab{b}.
\newblock \showarticletitle{Grow and Prune Compact, Fast, and Accurate LSTMs}.
\newblock \bibinfo{journal}{\emph{IEEE Trans. Comput.}} \bibinfo{volume}{69},
  \bibinfo{number}{3} (\bibinfo{year}{2019}), \bibinfo{pages}{441--452}.
\newblock


\bibitem[\protect\citeauthoryear{Dai, Yin, and Jha}{Dai et~al\mbox{.}}{2019c}]%
        {dai2019nest}
\bibfield{author}{\bibinfo{person}{Xiaoliang Dai}, \bibinfo{person}{Hongxu
  Yin}, {and} \bibinfo{person}{Niraj~K Jha}.} \bibinfo{year}{2019}\natexlab{c}.
\newblock \showarticletitle{NeST: A neural network synthesis tool based on a
  grow-and-prune paradigm}.
\newblock \bibinfo{journal}{\emph{IEEE Trans. Comput.}} \bibinfo{volume}{68},
  \bibinfo{number}{10} (\bibinfo{year}{2019}), \bibinfo{pages}{1487--1497}.
\newblock


\bibitem[\protect\citeauthoryear{Deng, Dong, Socher, Li, Li, and Fei-Fei}{Deng
  et~al\mbox{.}}{2009}]%
        {deng2009imagenet}
\bibfield{author}{\bibinfo{person}{Jia Deng}, \bibinfo{person}{Wei Dong},
  \bibinfo{person}{Richard Socher}, \bibinfo{person}{Li-Jia Li},
  \bibinfo{person}{Kai Li}, {and} \bibinfo{person}{Li Fei-Fei}.}
  \bibinfo{year}{2009}\natexlab{}.
\newblock \showarticletitle{Imagenet: A large-scale hierarchical image
  database}. In \bibinfo{booktitle}{\emph{Computer Vision and Pattern
  Recognition, 2009. CVPR 2009. IEEE Conference on}}. IEEE,
  \bibinfo{pages}{248--255}.
\newblock


\bibitem[\protect\citeauthoryear{Dong, Wang, Niu, Zhang, Lin, Li, Gong, Ren,
  Lin, and Tao}{Dong et~al\mbox{.}}{2020}]%
        {dong2020rtmobile}
\bibfield{author}{\bibinfo{person}{Peiyan Dong}, \bibinfo{person}{Siyue Wang},
  \bibinfo{person}{Wei Niu}, \bibinfo{person}{Chengming Zhang},
  \bibinfo{person}{Sheng Lin}, \bibinfo{person}{Zhengang Li},
  \bibinfo{person}{Yifan Gong}, \bibinfo{person}{Bin Ren}, \bibinfo{person}{Xue
  Lin}, {and} \bibinfo{person}{Dingwen Tao}.} \bibinfo{year}{2020}\natexlab{}.
\newblock \showarticletitle{Rtmobile: Beyond real-time mobile acceleration of
  rnns for speech recognition}. In \bibinfo{booktitle}{\emph{2020 57th ACM/IEEE
  Design Automation Conference (DAC)}}. IEEE, \bibinfo{pages}{1--6}.
\newblock


\bibitem[\protect\citeauthoryear{Dong and Yang}{Dong and Yang}{2019}]%
        {dong2019network}
\bibfield{author}{\bibinfo{person}{Xuanyi Dong} {and} \bibinfo{person}{Yi
  Yang}.} \bibinfo{year}{2019}\natexlab{}.
\newblock \showarticletitle{Network pruning via transformable architecture
  search}. In \bibinfo{booktitle}{\emph{Advances in Neural Information
  Processing Systems (NeurIPS)}}. \bibinfo{pages}{759--770}.
\newblock


\bibitem[\protect\citeauthoryear{Frankle and Carbin}{Frankle and
  Carbin}{2018}]%
        {frankle2018lottery}
\bibfield{author}{\bibinfo{person}{Jonathan Frankle} {and}
  \bibinfo{person}{Michael Carbin}.} \bibinfo{year}{2018}\natexlab{}.
\newblock \showarticletitle{The lottery ticket hypothesis: Finding sparse,
  trainable neural networks}. In \bibinfo{booktitle}{\emph{The International
  Conference on Learning Representations (ICLR)}}.
\newblock


\bibitem[\protect\citeauthoryear{Gong, Zhan, Li, Niu, Ma, Wang, Ren, Ding, Lin,
  Xu, et~al\mbox{.}}{Gong et~al\mbox{.}}{2020}]%
        {gong2020privacy}
\bibfield{author}{\bibinfo{person}{Yifan Gong}, \bibinfo{person}{Zheng Zhan},
  \bibinfo{person}{Zhengang Li}, \bibinfo{person}{Wei Niu},
  \bibinfo{person}{Xiaolong Ma}, \bibinfo{person}{Wenhao Wang},
  \bibinfo{person}{Bin Ren}, \bibinfo{person}{Caiwen Ding},
  \bibinfo{person}{Xue Lin}, \bibinfo{person}{Xiaolin Xu}, {et~al\mbox{.}}}
  \bibinfo{year}{2020}\natexlab{}.
\newblock \showarticletitle{A privacy-preserving-oriented dnn pruning and
  mobile acceleration framework}. In \bibinfo{booktitle}{\emph{Proceedings of
  the 2020 on Great Lakes Symposium on VLSI}}. \bibinfo{pages}{119--124}.
\newblock


\bibitem[\protect\citeauthoryear{Guo, Yao, and Chen}{Guo et~al\mbox{.}}{2016}]%
        {guo2016dynamic}
\bibfield{author}{\bibinfo{person}{Yiwen Guo}, \bibinfo{person}{Anbang Yao},
  {and} \bibinfo{person}{Yurong Chen}.} \bibinfo{year}{2016}\natexlab{}.
\newblock \showarticletitle{Dynamic network surgery for efficient dnns}. In
  \bibinfo{booktitle}{\emph{Advances in Neural Information Processing Systems
  (NeurIPS)}}.
\newblock


\bibitem[\protect\citeauthoryear{Han, Pool, Tran, and Dally}{Han
  et~al\mbox{.}}{2015}]%
        {han2015learning}
\bibfield{author}{\bibinfo{person}{Song Han}, \bibinfo{person}{Jeff Pool},
  \bibinfo{person}{John Tran}, {and} \bibinfo{person}{William Dally}.}
  \bibinfo{year}{2015}\natexlab{}.
\newblock \showarticletitle{Learning both weights and connections for efficient
  neural network}. In \bibinfo{booktitle}{\emph{Advances in neural information
  processing systems (NeurIPS)}}.
\newblock


\bibitem[\protect\citeauthoryear{Han, Shen, Philipose, Agarwal, Wolman, and
  Krishnamurthy}{Han et~al\mbox{.}}{2016}]%
        {han2016mcdnn}
\bibfield{author}{\bibinfo{person}{Seungyeop Han}, \bibinfo{person}{Haichen
  Shen}, \bibinfo{person}{Matthai Philipose}, \bibinfo{person}{Sharad Agarwal},
  \bibinfo{person}{Alec Wolman}, {and} \bibinfo{person}{Arvind Krishnamurthy}.}
  \bibinfo{year}{2016}\natexlab{}.
\newblock \showarticletitle{Mcdnn: An approximation-based execution framework
  for deep stream processing under resource constraints}. In
  \bibinfo{booktitle}{\emph{Proceedings of the 14th Annual International
  Conference on Mobile Systems, Applications, and Services (MobiSys)}}. ACM,
  \bibinfo{pages}{123--136}.
\newblock


\bibitem[\protect\citeauthoryear{He, Zhang, Ren, and Sun}{He
  et~al\mbox{.}}{2016}]%
        {he2016deep}
\bibfield{author}{\bibinfo{person}{Kaiming He}, \bibinfo{person}{Xiangyu
  Zhang}, \bibinfo{person}{Shaoqing Ren}, {and} \bibinfo{person}{Jian Sun}.}
  \bibinfo{year}{2016}\natexlab{}.
\newblock \showarticletitle{Deep residual learning for image recognition}. In
  \bibinfo{booktitle}{\emph{Proceedings of the IEEE Conference on Computer
  Vision and Pattern Recognition (CVPR)}}.
\newblock


\bibitem[\protect\citeauthoryear{He, Lin, Liu, Wang, Li, and Han}{He
  et~al\mbox{.}}{2018}]%
        {he2018amc}
\bibfield{author}{\bibinfo{person}{Yihui He}, \bibinfo{person}{Ji Lin},
  \bibinfo{person}{Zhijian Liu}, \bibinfo{person}{Hanrui Wang},
  \bibinfo{person}{Li-Jia Li}, {and} \bibinfo{person}{Song Han}.}
  \bibinfo{year}{2018}\natexlab{}.
\newblock \showarticletitle{Amc: Automl for model compression and acceleration
  on mobile devices}. In \bibinfo{booktitle}{\emph{Proceedings of the European
  Conference on Computer Vision (ECCV)}}.
\newblock


\bibitem[\protect\citeauthoryear{He, Liu, Wang, Hu, and Yang}{He
  et~al\mbox{.}}{2019}]%
        {he2019filter}
\bibfield{author}{\bibinfo{person}{Yang He}, \bibinfo{person}{Ping Liu},
  \bibinfo{person}{Ziwei Wang}, \bibinfo{person}{Zhilan Hu}, {and}
  \bibinfo{person}{Yi Yang}.} \bibinfo{year}{2019}\natexlab{}.
\newblock \showarticletitle{Filter pruning via geometric median for deep
  convolutional neural networks acceleration}. In
  \bibinfo{booktitle}{\emph{Proceedings of the IEEE Conference on Computer
  Vision and Pattern Recognition (CVPR)}}.
\newblock


\bibitem[\protect\citeauthoryear{He, Zhang, and Sun}{He et~al\mbox{.}}{2017}]%
        {he2017channel}
\bibfield{author}{\bibinfo{person}{Yihui He}, \bibinfo{person}{Xiangyu Zhang},
  {and} \bibinfo{person}{Jian Sun}.} \bibinfo{year}{2017}\natexlab{}.
\newblock \showarticletitle{Channel pruning for accelerating very deep neural
  networks}. In \bibinfo{booktitle}{\emph{Proceedings of the IEEE International
  Conference on Computer Vision (ICCV)}}.
\newblock


\bibitem[\protect\citeauthoryear{Hegde, Siddhartha, Ramasamy, and Kapre}{Hegde
  et~al\mbox{.}}{2016}]%
        {hegde2016caffepresso}
\bibfield{author}{\bibinfo{person}{Gopalakrishna Hegde},
  \bibinfo{person}{Siddhartha}, \bibinfo{person}{Nachiappan Ramasamy}, {and}
  \bibinfo{person}{Nachiket Kapre}.} \bibinfo{year}{2016}\natexlab{}.
\newblock \showarticletitle{CaffePresso: an optimized library for deep learning
  on embedded accelerator-based platforms}. In \bibinfo{booktitle}{\emph{2016
  International Conference on Compliers, Architectures, and Sythesis of
  Embedded Systems (CASES)}}. IEEE, \bibinfo{pages}{1--10}.
\newblock


\bibitem[\protect\citeauthoryear{Huynh, Lee, and Balan}{Huynh
  et~al\mbox{.}}{2017}]%
        {huynh2017deepmon}
\bibfield{author}{\bibinfo{person}{Loc~N Huynh}, \bibinfo{person}{Youngki Lee},
  {and} \bibinfo{person}{Rajesh~Krishna Balan}.}
  \bibinfo{year}{2017}\natexlab{}.
\newblock \showarticletitle{Deepmon: Mobile gpu-based deep learning framework
  for continuous vision applications}. In \bibinfo{booktitle}{\emph{Proceedings
  of the 15th Annual International Conference on Mobile Systems, Applications,
  and Services (MobiSys)}}. ACM, \bibinfo{pages}{82--95}.
\newblock


\bibitem[\protect\citeauthoryear{Jian, Gong, Zhan, Shi, Soltani, Wang, Dy,
  Chowdhury, Wang, and Ioannidis}{Jian et~al\mbox{.}}{2021}]%
        {jian2021radio}
\bibfield{author}{\bibinfo{person}{Tong Jian}, \bibinfo{person}{Yifan Gong},
  \bibinfo{person}{Zheng Zhan}, \bibinfo{person}{Runbin Shi},
  \bibinfo{person}{Nasim Soltani}, \bibinfo{person}{Zifeng Wang},
  \bibinfo{person}{Jennifer~G Dy}, \bibinfo{person}{Kaushik~Roy Chowdhury},
  \bibinfo{person}{Yanzhi Wang}, {and} \bibinfo{person}{Stratis Ioannidis}.}
  \bibinfo{year}{2021}\natexlab{}.
\newblock \showarticletitle{Radio Frequency Fingerprinting on the Edge}.
\newblock \bibinfo{journal}{\emph{IEEE Transactions on Mobile Computing}}
  (\bibinfo{year}{2021}).
\newblock


\bibitem[\protect\citeauthoryear{Klein, Falkner, Bartels, Hennig, and
  Hutter}{Klein et~al\mbox{.}}{2017}]%
        {klein2017fast}
\bibfield{author}{\bibinfo{person}{Aaron Klein}, \bibinfo{person}{Stefan
  Falkner}, \bibinfo{person}{Simon Bartels}, \bibinfo{person}{Philipp Hennig},
  {and} \bibinfo{person}{Frank Hutter}.} \bibinfo{year}{2017}\natexlab{}.
\newblock \showarticletitle{Fast bayesian optimization of machine learning
  hyperparameters on large datasets}. In \bibinfo{booktitle}{\emph{Artificial
  Intelligence and Statistics}}. PMLR, \bibinfo{pages}{528--536}.
\newblock


\bibitem[\protect\citeauthoryear{Lane, Bhattacharya, Georgiev, Forlivesi, Jiao,
  Qendro, and Kawsar}{Lane et~al\mbox{.}}{2016}]%
        {lane2016deepx}
\bibfield{author}{\bibinfo{person}{Nicholas~D Lane}, \bibinfo{person}{Sourav
  Bhattacharya}, \bibinfo{person}{Petko Georgiev}, \bibinfo{person}{Claudio
  Forlivesi}, \bibinfo{person}{Lei Jiao}, \bibinfo{person}{Lorena Qendro},
  {and} \bibinfo{person}{Fahim Kawsar}.} \bibinfo{year}{2016}\natexlab{}.
\newblock \showarticletitle{Deepx: A software accelerator for low-power deep
  learning inference on mobile devices}. In
  \bibinfo{booktitle}{\emph{Proceedings of the 15th International Conference on
  Information Processing in Sensor Networks}}. IEEE Press, \bibinfo{pages}{23}.
\newblock


\bibitem[\protect\citeauthoryear{Lane, Bhattacharya, Mathur, Georgiev,
  Forlivesi, and Kawsar}{Lane et~al\mbox{.}}{2017}]%
        {lane2017squeezing}
\bibfield{author}{\bibinfo{person}{Nicholas~D Lane}, \bibinfo{person}{Sourav
  Bhattacharya}, \bibinfo{person}{Akhil Mathur}, \bibinfo{person}{Petko
  Georgiev}, \bibinfo{person}{Claudio Forlivesi}, {and} \bibinfo{person}{Fahim
  Kawsar}.} \bibinfo{year}{2017}\natexlab{}.
\newblock \showarticletitle{Squeezing deep learning into mobile and embedded
  devices}.
\newblock \bibinfo{journal}{\emph{IEEE Pervasive Computing}}
  \bibinfo{volume}{16}, \bibinfo{number}{3} (\bibinfo{year}{2017}),
  \bibinfo{pages}{82--88}.
\newblock


\bibitem[\protect\citeauthoryear{Lane, Georgiev, and Qendro}{Lane
  et~al\mbox{.}}{2015}]%
        {lane2015deepear}
\bibfield{author}{\bibinfo{person}{Nicholas~D Lane}, \bibinfo{person}{Petko
  Georgiev}, {and} \bibinfo{person}{Lorena Qendro}.}
  \bibinfo{year}{2015}\natexlab{}.
\newblock \showarticletitle{DeepEar: robust smartphone audio sensing in
  unconstrained acoustic environments using deep learning}. In
  \bibinfo{booktitle}{\emph{Proceedings of the 2015 ACM International Joint
  Conference on Pervasive and Ubiquitous Computing}}. ACM,
  \bibinfo{pages}{283--294}.
\newblock


\bibitem[\protect\citeauthoryear{Li, Kadav, Durdanovic, Samet, and Graf}{Li
  et~al\mbox{.}}{2017}]%
        {li2017pruning}
\bibfield{author}{\bibinfo{person}{Hao Li}, \bibinfo{person}{Asim Kadav},
  \bibinfo{person}{Igor Durdanovic}, \bibinfo{person}{Hanan Samet}, {and}
  \bibinfo{person}{Hans~Peter Graf}.} \bibinfo{year}{2017}\natexlab{}.
\newblock \showarticletitle{Pruning filters for efficient convnets}.
\newblock \bibinfo{journal}{\emph{International Conference on Learning
  Representations (ICLR)}} (\bibinfo{year}{2017}).
\newblock


\bibitem[\protect\citeauthoryear{Li, Yuan, Niu, Cai, Sun, Li, Ren, Lin, and
  Wang}{Li et~al\mbox{.}}{2021}]%
        {li2021real}
\bibfield{author}{\bibinfo{person}{Hongjia Li}, \bibinfo{person}{Geng Yuan},
  \bibinfo{person}{Wei Niu}, \bibinfo{person}{Yuxuan Cai},
  \bibinfo{person}{Mengshu Sun}, \bibinfo{person}{Zhengang Li},
  \bibinfo{person}{Bin Ren}, \bibinfo{person}{Xue Lin}, {and}
  \bibinfo{person}{Yanzhi Wang}.} \bibinfo{year}{2021}\natexlab{}.
\newblock \showarticletitle{Real-Time Mobile Acceleration of DNNs: From
  Computer Vision to Medical Applications}. In \bibinfo{booktitle}{\emph{2021
  26th Asia and South Pacific Design Automation Conference (ASP-DAC)}}. IEEE,
  \bibinfo{pages}{581--586}.
\newblock


\bibitem[\protect\citeauthoryear{Li and Malik}{Li and Malik}{2016}]%
        {li2016learning}
\bibfield{author}{\bibinfo{person}{Ke Li} {and} \bibinfo{person}{Jitendra
  Malik}.} \bibinfo{year}{2016}\natexlab{}.
\newblock \showarticletitle{Learning to optimize}.
\newblock \bibinfo{journal}{\emph{arXiv preprint arXiv:1606.01885}}
  (\bibinfo{year}{2016}).
\newblock


\bibitem[\protect\citeauthoryear{Li, Wu, Yang, Fan, Zhang, and Liu}{Li
  et~al\mbox{.}}{2019}]%
        {li2019compressing}
\bibfield{author}{\bibinfo{person}{Tuanhui Li}, \bibinfo{person}{Baoyuan Wu},
  \bibinfo{person}{Yujiu Yang}, \bibinfo{person}{Yanbo Fan},
  \bibinfo{person}{Yong Zhang}, {and} \bibinfo{person}{Wei Liu}.}
  \bibinfo{year}{2019}\natexlab{}.
\newblock \showarticletitle{Compressing convolutional neural networks via
  factorized convolutional filters}. In \bibinfo{booktitle}{\emph{Proceedings
  of the IEEE Conference on Computer Vision and Pattern Recognition (CVPR)}}.
\newblock


\bibitem[\protect\citeauthoryear{Li, Dong, and Wang}{Li et~al\mbox{.}}{2020a}]%
        {li2019additive}
\bibfield{author}{\bibinfo{person}{Yuhang Li}, \bibinfo{person}{Xin Dong},
  {and} \bibinfo{person}{Wei Wang}.} \bibinfo{year}{2020}\natexlab{a}.
\newblock \showarticletitle{Additive Powers-of-Two Quantization: An Efficient
  Non-uniform Discretization for Neural Networks}. In
  \bibinfo{booktitle}{\emph{International Conference on Learning
  Representations (ICLR)}}.
\newblock


\bibitem[\protect\citeauthoryear{Li, Gong, Ma, Liu, Sun, Zhan, Kong, Yuan, and
  Wang}{Li et~al\mbox{.}}{2020b}]%
        {li2020ss}
\bibfield{author}{\bibinfo{person}{Zhengang Li}, \bibinfo{person}{Yifan Gong},
  \bibinfo{person}{Xiaolong Ma}, \bibinfo{person}{Sijia Liu},
  \bibinfo{person}{Mengshu Sun}, \bibinfo{person}{Zheng Zhan},
  \bibinfo{person}{Zhenglun Kong}, \bibinfo{person}{Geng Yuan}, {and}
  \bibinfo{person}{Yanzhi Wang}.} \bibinfo{year}{2020}\natexlab{b}.
\newblock \showarticletitle{SS-Auto: A Single-Shot, Automatic Structured Weight
  Pruning Framework of DNNs with Ultra-High Efficiency}.
\newblock \bibinfo{journal}{\emph{arXiv preprint arXiv:2001.08839}}
  (\bibinfo{year}{2020}).
\newblock


\bibitem[\protect\citeauthoryear{Lin, Maire, Belongie, Hays, Perona, Ramanan,
  Doll{\'a}r, and Zitnick}{Lin et~al\mbox{.}}{2014}]%
        {lin2014microsoft}
\bibfield{author}{\bibinfo{person}{Tsung-Yi Lin}, \bibinfo{person}{Michael
  Maire}, \bibinfo{person}{Serge Belongie}, \bibinfo{person}{James Hays},
  \bibinfo{person}{Pietro Perona}, \bibinfo{person}{Deva Ramanan},
  \bibinfo{person}{Piotr Doll{\'a}r}, {and} \bibinfo{person}{C~Lawrence
  Zitnick}.} \bibinfo{year}{2014}\natexlab{}.
\newblock \showarticletitle{Microsoft coco: Common objects in context}. In
  \bibinfo{booktitle}{\emph{European conference on computer vision}}. Springer,
  \bibinfo{pages}{740--755}.
\newblock


\bibitem[\protect\citeauthoryear{Liu, Ma, Xu, Wang, Tang, and Ye}{Liu
  et~al\mbox{.}}{2019a}]%
        {liu2019autocompress}
\bibfield{author}{\bibinfo{person}{Ning Liu}, \bibinfo{person}{Xiaolong Ma},
  \bibinfo{person}{Zhiyuan Xu}, \bibinfo{person}{Yanzhi Wang},
  \bibinfo{person}{Jian Tang}, {and} \bibinfo{person}{Jieping Ye}.}
  \bibinfo{year}{2019}\natexlab{a}.
\newblock \showarticletitle{AutoCompress: An Automatic DNN Structured Pruning
  Framework for Ultra-High Compression Rates}.
\newblock \bibinfo{journal}{\emph{arXiv preprint arXiv:1907.03141}}
  (\bibinfo{year}{2019}).
\newblock


\bibitem[\protect\citeauthoryear{Liu, Ma, Xu, Wang, Tang, and Ye}{Liu
  et~al\mbox{.}}{2019b}]%
        {liu2019autoslim}
\bibfield{author}{\bibinfo{person}{Ning Liu}, \bibinfo{person}{Xiaolong Ma},
  \bibinfo{person}{Zhiyuan Xu}, \bibinfo{person}{Yanzhi Wang},
  \bibinfo{person}{Jian Tang}, {and} \bibinfo{person}{Jieping Ye}.}
  \bibinfo{year}{2019}\natexlab{b}.
\newblock \showarticletitle{AutoSlim: An Automatic DNN Structured Pruning
  Framework for Ultra-High Compression Rates}.
\newblock \bibinfo{journal}{\emph{arXiv preprint arXiv:1907.03141}}
  (\bibinfo{year}{2019}).
\newblock


\bibitem[\protect\citeauthoryear{Liu, Ma, Xu, Wang, Tang, and Ye}{Liu
  et~al\mbox{.}}{2020}]%
        {Liu2020Autocompress}
\bibfield{author}{\bibinfo{person}{Ning Liu}, \bibinfo{person}{Xiaolong Ma},
  \bibinfo{person}{Zhiyuan Xu}, \bibinfo{person}{Yanzhi Wang},
  \bibinfo{person}{Jian Tang}, {and} \bibinfo{person}{Jieping Ye}.}
  \bibinfo{year}{2020}\natexlab{}.
\newblock \showarticletitle{AutoCompress: An Automatic DNN Structured Pruning
  Framework for Ultra-High Compression Rates}. In
  \bibinfo{booktitle}{\emph{AAAI}}.
\newblock


\bibitem[\protect\citeauthoryear{Liu, Li, Shen, Huang, Yan, and Zhang}{Liu
  et~al\mbox{.}}{2017}]%
        {liu2017learning}
\bibfield{author}{\bibinfo{person}{Zhuang Liu}, \bibinfo{person}{Jianguo Li},
  \bibinfo{person}{Zhiqiang Shen}, \bibinfo{person}{Gao Huang},
  \bibinfo{person}{Shoumeng Yan}, {and} \bibinfo{person}{Changshui Zhang}.}
  \bibinfo{year}{2017}\natexlab{}.
\newblock \showarticletitle{Learning efficient convolutional networks through
  network slimming}. In \bibinfo{booktitle}{\emph{Proceedings of the IEEE
  International Conference on Computer Vision (ICCV)}}.
\newblock


\bibitem[\protect\citeauthoryear{Liu, Mu, Zhang, Guo, Yang, Cheng, and Sun}{Liu
  et~al\mbox{.}}{2019c}]%
        {Liu_2019}
\bibfield{author}{\bibinfo{person}{Zechun Liu}, \bibinfo{person}{Haoyuan Mu},
  \bibinfo{person}{Xiangyu Zhang}, \bibinfo{person}{Zichao Guo},
  \bibinfo{person}{Xin Yang}, \bibinfo{person}{Kwang-Ting Cheng}, {and}
  \bibinfo{person}{Jian Sun}.} \bibinfo{year}{2019}\natexlab{c}.
\newblock \showarticletitle{MetaPruning: Meta Learning for Automatic Neural
  Network Channel Pruning}.
\newblock \bibinfo{journal}{\emph{2019 IEEE/CVF International Conference on
  Computer Vision (ICCV)}} (\bibinfo{date}{Oct} \bibinfo{year}{2019}).
\newblock
\showISBNx{9781728148038}
\urldef\tempurl%
\url{https://doi.org/10.1109/iccv.2019.00339}
\showDOI{\tempurl}


\bibitem[\protect\citeauthoryear{Liu, Sun, Zhou, Huang, and Darrell}{Liu
  et~al\mbox{.}}{2018}]%
        {liu2018rethinking}
\bibfield{author}{\bibinfo{person}{Zhuang Liu}, \bibinfo{person}{Mingjie Sun},
  \bibinfo{person}{Tinghui Zhou}, \bibinfo{person}{Gao Huang}, {and}
  \bibinfo{person}{Trevor Darrell}.} \bibinfo{year}{2018}\natexlab{}.
\newblock \showarticletitle{Rethinking the Value of Network Pruning}. In
  \bibinfo{booktitle}{\emph{International Conference on Learning
  Representations}}.
\newblock


\bibitem[\protect\citeauthoryear{Luo, Wu, and Lin}{Luo et~al\mbox{.}}{2017}]%
        {luo2017thinet}
\bibfield{author}{\bibinfo{person}{Jian-Hao Luo}, \bibinfo{person}{Jianxin Wu},
  {and} \bibinfo{person}{Weiyao Lin}.} \bibinfo{year}{2017}\natexlab{}.
\newblock \showarticletitle{Thinet: A filter level pruning method for deep
  neural network compression}. In \bibinfo{booktitle}{\emph{Proceedings of the
  IEEE International Conference on Computer Vision (ICCV)}}.
\newblock


\bibitem[\protect\citeauthoryear{Ma, Guo, Niu, Lin, Tang, Ma, Ren, and Wang}{Ma
  et~al\mbox{.}}{2020a}]%
        {ma2020pconv}
\bibfield{author}{\bibinfo{person}{Xiaolong Ma}, \bibinfo{person}{Fu-Ming Guo},
  \bibinfo{person}{Wei Niu}, \bibinfo{person}{Xue Lin}, \bibinfo{person}{Jian
  Tang}, \bibinfo{person}{Kaisheng Ma}, \bibinfo{person}{Bin Ren}, {and}
  \bibinfo{person}{Yanzhi Wang}.} \bibinfo{year}{2020}\natexlab{a}.
\newblock \showarticletitle{Pconv: The missing but desirable sparsity in dnn
  weight pruning for real-time execution on mobile devices}. In
  \bibinfo{booktitle}{\emph{Thirty-Fourth AAAI conference on artificial
  intelligence (AAAI)}}.
\newblock


\bibitem[\protect\citeauthoryear{Ma, Li, Gong, Zhang, Niu, Zhan, Zhao, Tang,
  Lin, Ren, et~al\mbox{.}}{Ma et~al\mbox{.}}{2020b}]%
        {ma2020blk}
\bibfield{author}{\bibinfo{person}{Xiaolong Ma}, \bibinfo{person}{Zhengang Li},
  \bibinfo{person}{Yifan Gong}, \bibinfo{person}{Tianyun Zhang},
  \bibinfo{person}{Wei Niu}, \bibinfo{person}{Zheng Zhan}, \bibinfo{person}{Pu
  Zhao}, \bibinfo{person}{Jian Tang}, \bibinfo{person}{Xue Lin},
  \bibinfo{person}{Bin Ren}, {et~al\mbox{.}}} \bibinfo{year}{2020}\natexlab{b}.
\newblock \showarticletitle{Blk-rew: A unified block-based dnn pruning
  framework using reweighted regularization method}.
\newblock \bibinfo{journal}{\emph{arXiv preprint arXiv:2001.08357}}
  (\bibinfo{year}{2020}).
\newblock


\bibitem[\protect\citeauthoryear{Ma, Lin, Ye, He, Zhang, Yuan, Tan, Li, Fan,
  Qian, Lin, Ma, and Wang}{Ma et~al\mbox{.}}{2019a}]%
        {ma2019nonstructured}
\bibfield{author}{\bibinfo{person}{Xiaolong Ma}, \bibinfo{person}{Sheng Lin},
  \bibinfo{person}{Shaokai Ye}, \bibinfo{person}{Zhezhi He},
  \bibinfo{person}{Linfeng Zhang}, \bibinfo{person}{Geng Yuan},
  \bibinfo{person}{Sia~Huat Tan}, \bibinfo{person}{Zhengang Li},
  \bibinfo{person}{Deliang Fan}, \bibinfo{person}{Xuehai Qian},
  \bibinfo{person}{Xue Lin}, \bibinfo{person}{Kaisheng Ma}, {and}
  \bibinfo{person}{Yanzhi Wang}.} \bibinfo{year}{2019}\natexlab{a}.
\newblock \bibinfo{title}{Non-Structured DNN Weight Pruning -- Is It Beneficial
  in Any Platform?}
\newblock
\newblock
\showeprint[arxiv]{1907.02124}~[cs.LG]


\bibitem[\protect\citeauthoryear{Ma, Niu, Zhang, Liu, Lin, Li, Wen, Chen, Tang,
  Ma, et~al\mbox{.}}{Ma et~al\mbox{.}}{2020c}]%
        {ma2020image}
\bibfield{author}{\bibinfo{person}{Xiaolong Ma}, \bibinfo{person}{Wei Niu},
  \bibinfo{person}{Tianyun Zhang}, \bibinfo{person}{Sijia Liu},
  \bibinfo{person}{Sheng Lin}, \bibinfo{person}{Hongjia Li},
  \bibinfo{person}{Wujie Wen}, \bibinfo{person}{Xiang Chen},
  \bibinfo{person}{Jian Tang}, \bibinfo{person}{Kaisheng Ma}, {et~al\mbox{.}}}
  \bibinfo{year}{2020}\natexlab{c}.
\newblock \showarticletitle{An image enhancing pattern-based sparsity for
  real-time inference on mobile devices}. In \bibinfo{booktitle}{\emph{European
  Conference on Computer Vision}}. Springer, \bibinfo{pages}{629--645}.
\newblock


\bibitem[\protect\citeauthoryear{Ma, Yuan, Lin, Ding, Yu, Liu, Wen, Chen, and
  Wang}{Ma et~al\mbox{.}}{2020d}]%
        {ma2019tiny}
\bibfield{author}{\bibinfo{person}{Xiaolong Ma}, \bibinfo{person}{Geng Yuan},
  \bibinfo{person}{Sheng Lin}, \bibinfo{person}{Caiwen Ding},
  \bibinfo{person}{Fuxun Yu}, \bibinfo{person}{Tao Liu}, \bibinfo{person}{Wujie
  Wen}, \bibinfo{person}{Xiang Chen}, {and} \bibinfo{person}{Yanzhi Wang}.}
  \bibinfo{year}{2020}\natexlab{d}.
\newblock \showarticletitle{Tiny but Accurate: A Pruned, Quantized and
  Optimized Memristor Crossbar Framework for Ultra Efficient DNN
  Implementation}. In \bibinfo{booktitle}{\emph{ASP-DAC}}.
\newblock


\bibitem[\protect\citeauthoryear{Ma, Yuan, Lin, Ding, Yu, Liu, Wen, Chen, and
  Wang}{Ma et~al\mbox{.}}{2020e}]%
        {ma2020tiny}
\bibfield{author}{\bibinfo{person}{Xiaolong Ma}, \bibinfo{person}{Geng Yuan},
  \bibinfo{person}{Sheng Lin}, \bibinfo{person}{Caiwen Ding},
  \bibinfo{person}{Fuxun Yu}, \bibinfo{person}{Tao Liu}, \bibinfo{person}{Wujie
  Wen}, \bibinfo{person}{Xiang Chen}, {and} \bibinfo{person}{Yanzhi Wang}.}
  \bibinfo{year}{2020}\natexlab{e}.
\newblock \showarticletitle{Tiny but accurate: A pruned, quantized and
  optimized memristor crossbar framework for ultra efficient dnn
  implementation}. In \bibinfo{booktitle}{\emph{2020 25th Asia and South
  Pacific Design Automation Conference (ASP-DAC)}}. IEEE,
  \bibinfo{pages}{301--306}.
\newblock


\bibitem[\protect\citeauthoryear{Ma, Yuan, Lin, Li, Sun, and Wang}{Ma
  et~al\mbox{.}}{2019b}]%
        {ma2019resnet}
\bibfield{author}{\bibinfo{person}{Xiaolong Ma}, \bibinfo{person}{Geng Yuan},
  \bibinfo{person}{Sheng Lin}, \bibinfo{person}{Zhengang Li},
  \bibinfo{person}{Hao Sun}, {and} \bibinfo{person}{Yanzhi Wang}.}
  \bibinfo{year}{2019}\natexlab{b}.
\newblock \showarticletitle{ResNet Can Be Pruned 60$\times$: Introducing
  Network Purification and Unused Path Removal (P-RM) after Weight Pruning}. In
  \bibinfo{booktitle}{\emph{2019 IEEE/ACM International Symposium on Nanoscale
  Architectures (NANOARCH)}}. IEEE, \bibinfo{pages}{1--2}.
\newblock


\bibitem[\protect\citeauthoryear{Min, Wang, Chen, Xu, and Chen}{Min
  et~al\mbox{.}}{2018}]%
        {min20182pfpce}
\bibfield{author}{\bibinfo{person}{Chuhan Min}, \bibinfo{person}{Aosen Wang},
  \bibinfo{person}{Yiran Chen}, \bibinfo{person}{Wenyao Xu}, {and}
  \bibinfo{person}{Xin Chen}.} \bibinfo{year}{2018}\natexlab{}.
\newblock \showarticletitle{2pfpce: Two-phase filter pruning based on
  conditional entropy}.
\newblock \bibinfo{journal}{\emph{arXiv preprint arXiv:1809.02220}}
  (\bibinfo{year}{2018}).
\newblock


\bibitem[\protect\citeauthoryear{Niu, Kong, Yuan, Jiang, Guan, Ding, Zhao, Liu,
  Ren, and Wang}{Niu et~al\mbox{.}}{2020a}]%
        {niu2020achieving}
\bibfield{author}{\bibinfo{person}{Wei Niu}, \bibinfo{person}{Zhenglun Kong},
  \bibinfo{person}{Geng Yuan}, \bibinfo{person}{Weiwen Jiang},
  \bibinfo{person}{Jiexiong Guan}, \bibinfo{person}{Caiwen Ding},
  \bibinfo{person}{Pu Zhao}, \bibinfo{person}{Sijia Liu}, \bibinfo{person}{Bin
  Ren}, {and} \bibinfo{person}{Yanzhi Wang}.} \bibinfo{year}{2020}\natexlab{a}.
\newblock \showarticletitle{Achieving Real-Time Execution of Transformer-based
  Large-scale Models on Mobile with Compiler-aware Neural Architecture
  Optimization}.
\newblock \bibinfo{journal}{\emph{arXiv preprint arXiv:2009.06823}}
  (\bibinfo{year}{2020}).
\newblock


\bibitem[\protect\citeauthoryear{Niu, Ma, Lin, Wang, Qian, Lin, Wang, and
  Ren}{Niu et~al\mbox{.}}{2020b}]%
        {niu2020patdnn}
\bibfield{author}{\bibinfo{person}{Wei Niu}, \bibinfo{person}{Xiaolong Ma},
  \bibinfo{person}{Sheng Lin}, \bibinfo{person}{Shihao Wang},
  \bibinfo{person}{Xuehai Qian}, \bibinfo{person}{Xue Lin},
  \bibinfo{person}{Yanzhi Wang}, {and} \bibinfo{person}{Bin Ren}.}
  \bibinfo{year}{2020}\natexlab{b}.
\newblock \showarticletitle{Patdnn: Achieving real-time DNN execution on mobile
  devices with pattern-based weight pruning}. In
  \bibinfo{booktitle}{\emph{Proceedings of the Twenty-Fifth International
  Conference on Architectural Support for Programming Languages and Operating
  Systems (ASPLOS)}}.
\newblock


\bibitem[\protect\citeauthoryear{Ota, Dao, Mezaris, and Natale}{Ota
  et~al\mbox{.}}{2017}]%
        {ota2017deep}
\bibfield{author}{\bibinfo{person}{Kaoru Ota}, \bibinfo{person}{Minh~Son Dao},
  \bibinfo{person}{Vasileios Mezaris}, {and} \bibinfo{person}{Francesco GB~De
  Natale}.} \bibinfo{year}{2017}\natexlab{}.
\newblock \showarticletitle{Deep learning for mobile multimedia: A survey}.
\newblock \bibinfo{journal}{\emph{ACM Transactions on Multimedia Computing,
  Communications, and Applications (TOMM)}} \bibinfo{volume}{13},
  \bibinfo{number}{3s} (\bibinfo{year}{2017}), \bibinfo{pages}{1--22}.
\newblock


\bibitem[\protect\citeauthoryear{Ren, Zhang, Ye, Li, Xu, Qian, Lin, and
  Wang}{Ren et~al\mbox{.}}{2019a}]%
        {ren2019admm}
\bibfield{author}{\bibinfo{person}{Ao Ren}, \bibinfo{person}{Tianyun Zhang},
  \bibinfo{person}{Shaokai Ye}, \bibinfo{person}{Jiayu Li},
  \bibinfo{person}{Wenyao Xu}, \bibinfo{person}{Xuehai Qian},
  \bibinfo{person}{Xue Lin}, {and} \bibinfo{person}{Yanzhi Wang}.}
  \bibinfo{year}{2019}\natexlab{a}.
\newblock \showarticletitle{Admm-nn: An algorithm-hardware co-design framework
  of dnns using alternating direction methods of multipliers}. In
  \bibinfo{booktitle}{\emph{Proceedings of the Twenty-Fourth International
  Conference on Architectural Support for Programming Languages and Operating
  Systems (ASPLOS)}}.
\newblock


\bibitem[\protect\citeauthoryear{Ren, Zhang, Ye, Xu, Qian, Lin, and Wang}{Ren
  et~al\mbox{.}}{2019b}]%
        {ren2019ADMMNN}
\bibfield{author}{\bibinfo{person}{Ao Ren}, \bibinfo{person}{Tianyun Zhang},
  \bibinfo{person}{Shaokai Ye}, \bibinfo{person}{Wenyao Xu},
  \bibinfo{person}{Xuehai Qian}, \bibinfo{person}{Xue Lin}, {and}
  \bibinfo{person}{Yanzhi Wang}.} \bibinfo{year}{2019}\natexlab{b}.
\newblock \showarticletitle{ADMM-NN: an algorithm-hardware co-design framework
  of DNNs using alternating direction methods of multipliers}. In
  \bibinfo{booktitle}{\emph{ASPLOS}}.
\newblock


\bibitem[\protect\citeauthoryear{Sandler, Howard, Zhu, Zhmoginov, and
  Chen}{Sandler et~al\mbox{.}}{2018}]%
        {sandler2018mobilenetv2}
\bibfield{author}{\bibinfo{person}{Mark Sandler}, \bibinfo{person}{Andrew
  Howard}, \bibinfo{person}{Menglong Zhu}, \bibinfo{person}{Andrey Zhmoginov},
  {and} \bibinfo{person}{Liang-Chieh Chen}.} \bibinfo{year}{2018}\natexlab{}.
\newblock \showarticletitle{Mobilenetv2: Inverted residuals and linear
  bottlenecks}. In \bibinfo{booktitle}{\emph{Proceedings of the IEEE Conference
  on Computer Vision and Pattern Recognition (CVPR)}}.
\newblock


\bibitem[\protect\citeauthoryear{Simonyan and Zisserman}{Simonyan and
  Zisserman}{2014}]%
        {simonyan2014very}
\bibfield{author}{\bibinfo{person}{Karen Simonyan} {and}
  \bibinfo{person}{Andrew Zisserman}.} \bibinfo{year}{2014}\natexlab{}.
\newblock \showarticletitle{Very deep convolutional networks for large-scale
  image recognition}.
\newblock \bibinfo{journal}{\emph{arXiv:1409.1556}} (\bibinfo{year}{2014}).
\newblock


\bibitem[\protect\citeauthoryear{Sutton and Barto}{Sutton and Barto}{2018}]%
        {sutton2018reinforcement}
\bibfield{author}{\bibinfo{person}{Richard~S Sutton} {and}
  \bibinfo{person}{Andrew~G Barto}.} \bibinfo{year}{2018}\natexlab{}.
\newblock \bibinfo{booktitle}{\emph{Reinforcement learning: An introduction}}.
\newblock \bibinfo{publisher}{MIT press}.
\newblock


\bibitem[\protect\citeauthoryear{Sutton, McAllester, Singh, Mansour,
  et~al\mbox{.}}{Sutton et~al\mbox{.}}{1999}]%
        {sutton1999policy}
\bibfield{author}{\bibinfo{person}{Richard~S Sutton}, \bibinfo{person}{David~A
  McAllester}, \bibinfo{person}{Satinder~P Singh}, \bibinfo{person}{Yishay
  Mansour}, {et~al\mbox{.}}} \bibinfo{year}{1999}\natexlab{}.
\newblock \showarticletitle{Policy gradient methods for reinforcement learning
  with function approximation.}. In \bibinfo{booktitle}{\emph{NIPs}},
  Vol.~\bibinfo{volume}{99}. Citeseer, \bibinfo{pages}{1057--1063}.
\newblock


\bibitem[\protect\citeauthoryear{Tan, Chen, Pang, Vasudevan, Sandler, Howard,
  and Le}{Tan et~al\mbox{.}}{2019}]%
        {tan2019mnasnet}
\bibfield{author}{\bibinfo{person}{Mingxing Tan}, \bibinfo{person}{Bo Chen},
  \bibinfo{person}{Ruoming Pang}, \bibinfo{person}{Vijay Vasudevan},
  \bibinfo{person}{Mark Sandler}, \bibinfo{person}{Andrew Howard}, {and}
  \bibinfo{person}{Quoc~V Le}.} \bibinfo{year}{2019}\natexlab{}.
\newblock \showarticletitle{Mnasnet: Platform-aware neural architecture search
  for mobile}. In \bibinfo{booktitle}{\emph{Proceedings of the IEEE Conference
  on Computer Vision and Pattern Recognition (CVPR)}}.
  \bibinfo{pages}{2820--2828}.
\newblock


\bibitem[\protect\citeauthoryear{Wang, Ye, He, Ma, Zhang, Lin, Yuan, Tan, Li,
  Fan, Qian, Lin, and Ma}{Wang et~al\mbox{.}}{2019}]%
        {Wang2019NonstructuredDW}
\bibfield{author}{\bibinfo{person}{Yanzhi Wang}, \bibinfo{person}{Shaokai Ye},
  \bibinfo{person}{Zhezhi He}, \bibinfo{person}{Xiaolong Ma},
  \bibinfo{person}{Linfeng Zhang}, \bibinfo{person}{Sheng Lin},
  \bibinfo{person}{Geng Yuan}, \bibinfo{person}{Sia~Huat Tan},
  \bibinfo{person}{Zhengang Li}, \bibinfo{person}{Deliang Fan},
  \bibinfo{person}{Xuehai Qian}, \bibinfo{person}{Xue Lin}, {and}
  \bibinfo{person}{Kaisheng Ma}.} \bibinfo{year}{2019}\natexlab{}.
\newblock \showarticletitle{Non-structured DNN weight pruning considered
  harmful}.
\newblock \bibinfo{journal}{\emph{arXiv:1907.02124}} (\bibinfo{year}{2019}).
\newblock


\bibitem[\protect\citeauthoryear{Wen, Wu, Wang, Chen, and Li}{Wen
  et~al\mbox{.}}{2016}]%
        {wen2016learning}
\bibfield{author}{\bibinfo{person}{Wei Wen}, \bibinfo{person}{Chunpeng Wu},
  \bibinfo{person}{Yandan Wang}, \bibinfo{person}{Yiran Chen}, {and}
  \bibinfo{person}{Hai Li}.} \bibinfo{year}{2016}\natexlab{}.
\newblock \showarticletitle{Learning structured sparsity in deep neural
  networks}. In \bibinfo{booktitle}{\emph{Advances in Neural Information
  Processing Systems (NeurIPS)}}.
\newblock


\bibitem[\protect\citeauthoryear{Wu, Dai, Zhang, Wang, Sun, Wu, Tian, Vajda,
  Jia, and Keutzer}{Wu et~al\mbox{.}}{2019}]%
        {wu2019fbnet}
\bibfield{author}{\bibinfo{person}{Bichen Wu}, \bibinfo{person}{Xiaoliang Dai},
  \bibinfo{person}{Peizhao Zhang}, \bibinfo{person}{Yanghan Wang},
  \bibinfo{person}{Fei Sun}, \bibinfo{person}{Yiming Wu},
  \bibinfo{person}{Yuandong Tian}, \bibinfo{person}{Peter Vajda},
  \bibinfo{person}{Yangqing Jia}, {and} \bibinfo{person}{Kurt Keutzer}.}
  \bibinfo{year}{2019}\natexlab{}.
\newblock \showarticletitle{Fbnet: Hardware-aware efficient convnet design via
  differentiable neural architecture search}. In
  \bibinfo{booktitle}{\emph{Proceedings of the IEEE Conference on Computer
  Vision and Pattern Recognition (CVPR)}}. \bibinfo{pages}{10734--10742}.
\newblock


\bibitem[\protect\citeauthoryear{Xu, Zhu, Liu, Lin, and Liu}{Xu
  et~al\mbox{.}}{2018}]%
        {xu2018deepcache}
\bibfield{author}{\bibinfo{person}{Mengwei Xu}, \bibinfo{person}{Mengze Zhu},
  \bibinfo{person}{Yunxin Liu}, \bibinfo{person}{Felix~Xiaozhu Lin}, {and}
  \bibinfo{person}{Xuanzhe Liu}.} \bibinfo{year}{2018}\natexlab{}.
\newblock \showarticletitle{DeepCache: Principled Cache for Mobile Deep
  Vision}. In \bibinfo{booktitle}{\emph{Proceedings of the 24th Annual
  International Conference on Mobile Computing and Networking}}. ACM,
  \bibinfo{pages}{129--144}.
\newblock


\bibitem[\protect\citeauthoryear{Yang, Howard, Chen, Zhang, Go, Sandler, Sze,
  and Adam}{Yang et~al\mbox{.}}{2018}]%
        {Yang_2018}
\bibfield{author}{\bibinfo{person}{Tien-Ju Yang}, \bibinfo{person}{Andrew
  Howard}, \bibinfo{person}{Bo Chen}, \bibinfo{person}{Xiao Zhang},
  \bibinfo{person}{Alec Go}, \bibinfo{person}{Mark Sandler},
  \bibinfo{person}{Vivienne Sze}, {and} \bibinfo{person}{Hartwig Adam}.}
  \bibinfo{year}{2018}\natexlab{}.
\newblock \showarticletitle{NetAdapt: Platform-Aware Neural Network Adaptation
  for Mobile Applications}.
\newblock \bibinfo{journal}{\emph{Lecture Notes in Computer Science}}
  (\bibinfo{year}{2018}), \bibinfo{pages}{289–304}.
\newblock
\showISBNx{9783030012496}
\showISSN{1611-3349}
\urldef\tempurl%
\url{https://doi.org/10.1007/978-3-030-01249-6_18}
\showDOI{\tempurl}


\bibitem[\protect\citeauthoryear{Yao, Hu, Zhao, Zhang, and Abdelzaher}{Yao
  et~al\mbox{.}}{2017}]%
        {yao2017deepsense}
\bibfield{author}{\bibinfo{person}{Shuochao Yao}, \bibinfo{person}{Shaohan Hu},
  \bibinfo{person}{Yiran Zhao}, \bibinfo{person}{Aston Zhang}, {and}
  \bibinfo{person}{Tarek Abdelzaher}.} \bibinfo{year}{2017}\natexlab{}.
\newblock \showarticletitle{Deepsense: A unified deep learning framework for
  time-series mobile sensing data processing}. In
  \bibinfo{booktitle}{\emph{Proceedings of the 26th International Conference on
  World Wide Web}}. \bibinfo{pages}{351--360}.
\newblock


\bibitem[\protect\citeauthoryear{Yu, Li, Chen, Lai, Morariu, Han, Gao, Lin, and
  Davis}{Yu et~al\mbox{.}}{2018}]%
        {yu2018nisp}
\bibfield{author}{\bibinfo{person}{Ruichi Yu}, \bibinfo{person}{Ang Li},
  \bibinfo{person}{Chun-Fu Chen}, \bibinfo{person}{Jui-Hsin Lai},
  \bibinfo{person}{Vlad~I Morariu}, \bibinfo{person}{Xintong Han},
  \bibinfo{person}{Mingfei Gao}, \bibinfo{person}{Ching-Yung Lin}, {and}
  \bibinfo{person}{Larry~S Davis}.} \bibinfo{year}{2018}\natexlab{}.
\newblock \showarticletitle{Nisp: Pruning networks using neuron importance
  score propagation}. In \bibinfo{booktitle}{\emph{Proceedings of the IEEE
  Conference on Computer Vision and Pattern Recognition (CVPR)}}.
\newblock


\bibitem[\protect\citeauthoryear{Yuan, Behnam, Li, Shafiee, Lin, Ma, Liu, Qian,
  Bojnordi, Wang, and et~al.}{Yuan et~al\mbox{.}}{2021a}]%
        {yuan2021forms}
\bibfield{author}{\bibinfo{person}{Geng Yuan}, \bibinfo{person}{Payman Behnam},
  \bibinfo{person}{Zhengang Li}, \bibinfo{person}{Ali Shafiee},
  \bibinfo{person}{Sheng Lin}, \bibinfo{person}{Xiaolong Ma},
  \bibinfo{person}{Hang Liu}, \bibinfo{person}{Xuehai Qian},
  \bibinfo{person}{Mahdi~Nazm Bojnordi}, \bibinfo{person}{Yanzhi Wang}, {and}
  \bibinfo{person}{et al.}} \bibinfo{year}{2021}\natexlab{a}.
\newblock \showarticletitle{FORMS: Fine-grained Polarized ReRAM-based In-situ
  Computation for Mixed-signal DNN Accelerator}.
\newblock \bibinfo{journal}{\emph{2021 ACM/IEEE 48th Annual International
  Symposium on Computer Architecture (ISCA)}} (\bibinfo{date}{Jun}
  \bibinfo{year}{2021}).
\newblock
\urldef\tempurl%
\url{https://doi.org/10.1109/isca52012.2021.00029}
\showDOI{\tempurl}


\bibitem[\protect\citeauthoryear{Yuan, Ma, Ding, Lin, Zhang, Jalali, Zhao,
  Jiang, Soundarajan, and Wang}{Yuan et~al\mbox{.}}{2019a}]%
        {yuan2019ultra}
\bibfield{author}{\bibinfo{person}{Geng Yuan}, \bibinfo{person}{Xiaolong Ma},
  \bibinfo{person}{Caiwen Ding}, \bibinfo{person}{Sheng Lin},
  \bibinfo{person}{Tianyun Zhang}, \bibinfo{person}{Zeinab~S Jalali},
  \bibinfo{person}{Yilong Zhao}, \bibinfo{person}{Li Jiang},
  \bibinfo{person}{Sucheta Soundarajan}, {and} \bibinfo{person}{Yanzhi Wang}.}
  \bibinfo{year}{2019}\natexlab{a}.
\newblock \showarticletitle{An ultra-efficient memristor-based dnn framework
  with structured weight pruning and quantization using admm}. In
  \bibinfo{booktitle}{\emph{2019 IEEE/ACM International Symposium on Low Power
  Electronics and Design (ISLPED)}}. IEEE, \bibinfo{pages}{1--6}.
\newblock


\bibitem[\protect\citeauthoryear{Yuan, Ma, Lin, Li, and Ding}{Yuan
  et~al\mbox{.}}{2019b}]%
        {yuan2019sot}
\bibfield{author}{\bibinfo{person}{Geng Yuan}, \bibinfo{person}{Xiaolong Ma},
  \bibinfo{person}{Sheng Lin}, \bibinfo{person}{Zhengang Li}, {and}
  \bibinfo{person}{Caiwen Ding}.} \bibinfo{year}{2019}\natexlab{b}.
\newblock \showarticletitle{A SOT-MRAM-based Processing-In-Memory Engine for
  Highly Compressed DNN Implementation}.
\newblock \bibinfo{journal}{\emph{arXiv preprint arXiv:1912.05416}}
  (\bibinfo{year}{2019}).
\newblock


\bibitem[\protect\citeauthoryear{Yuan, Ma, Niu, Li, Kong, Liu, Gong, Zhan, He,
  Jin, Wang, Qin, Ren, Wang, Liu, and Lin}{Yuan et~al\mbox{.}}{2021b}]%
        {yuan2021mest}
\bibfield{author}{\bibinfo{person}{Geng Yuan}, \bibinfo{person}{Xiaolong Ma},
  \bibinfo{person}{Wei Niu}, \bibinfo{person}{Zhengang Li},
  \bibinfo{person}{Zhenglun Kong}, \bibinfo{person}{Ning Liu},
  \bibinfo{person}{Yifan Gong}, \bibinfo{person}{Zheng Zhan},
  \bibinfo{person}{Chaoyang He}, \bibinfo{person}{Qing Jin},
  \bibinfo{person}{Siyue Wang}, \bibinfo{person}{Minghai Qin},
  \bibinfo{person}{Bin Ren}, \bibinfo{person}{Yanzhi Wang},
  \bibinfo{person}{Sijia Liu}, {and} \bibinfo{person}{Xue Lin}.}
  \bibinfo{year}{2021}\natexlab{b}.
\newblock \bibinfo{title}{MEST: Accurate and Fast Memory-Economic Sparse
  Training Framework on the Edge}.
\newblock
\newblock
\showeprint[arxiv]{2110.14032}~[cs.LG]


\bibitem[\protect\citeauthoryear{Zhan, Gong, Zhao, Yuan, Niu, Wu, Zhang,
  Jayaweera, Kaeli, Ren, et~al\mbox{.}}{Zhan et~al\mbox{.}}{2021}]%
        {zhan2021achieving}
\bibfield{author}{\bibinfo{person}{Zheng Zhan}, \bibinfo{person}{Yifan Gong},
  \bibinfo{person}{Pu Zhao}, \bibinfo{person}{Geng Yuan}, \bibinfo{person}{Wei
  Niu}, \bibinfo{person}{Yushu Wu}, \bibinfo{person}{Tianyun Zhang},
  \bibinfo{person}{Malith Jayaweera}, \bibinfo{person}{David Kaeli},
  \bibinfo{person}{Bin Ren}, {et~al\mbox{.}}} \bibinfo{year}{2021}\natexlab{}.
\newblock \showarticletitle{Achieving on-Mobile Real-Time Super-Resolution with
  Neural Architecture and Pruning Search}. In
  \bibinfo{booktitle}{\emph{Proceedings of the IEEE/CVF International
  Conference on Computer Vision}}. \bibinfo{pages}{4821--4831}.
\newblock


\bibitem[\protect\citeauthoryear{Zhang, Patras, and Haddadi}{Zhang
  et~al\mbox{.}}{2019}]%
        {zhang2019deep}
\bibfield{author}{\bibinfo{person}{Chaoyun Zhang}, \bibinfo{person}{Paul
  Patras}, {and} \bibinfo{person}{Hamed Haddadi}.}
  \bibinfo{year}{2019}\natexlab{}.
\newblock \showarticletitle{Deep learning in mobile and wireless networking: A
  survey}.
\newblock \bibinfo{journal}{\emph{IEEE Communications Surveys \& Tutorials}}
  \bibinfo{volume}{21}, \bibinfo{number}{3} (\bibinfo{year}{2019}),
  \bibinfo{pages}{2224--2287}.
\newblock


\bibitem[\protect\citeauthoryear{Zhang, Ye, Zhang, Tang, Wen, Fardad, and
  Wang}{Zhang et~al\mbox{.}}{2018a}]%
        {zhang2018systematic}
\bibfield{author}{\bibinfo{person}{Tianyun Zhang}, \bibinfo{person}{Shaokai
  Ye}, \bibinfo{person}{Kaiqi Zhang}, \bibinfo{person}{Jian Tang},
  \bibinfo{person}{Wujie Wen}, \bibinfo{person}{Makan Fardad}, {and}
  \bibinfo{person}{Yanzhi Wang}.} \bibinfo{year}{2018}\natexlab{a}.
\newblock \showarticletitle{A systematic dnn weight pruning framework using
  alternating direction method of multipliers}. In
  \bibinfo{booktitle}{\emph{Proceedings of the European Conference on Computer
  Vision (ECCV)}}.
\newblock


\bibitem[\protect\citeauthoryear{Zhang, Zhang, Ye, Li, Tang, Wen, Lin, Fardad,
  and Wang}{Zhang et~al\mbox{.}}{2018b}]%
        {zhang2018adam}
\bibfield{author}{\bibinfo{person}{Tianyun Zhang}, \bibinfo{person}{Kaiqi
  Zhang}, \bibinfo{person}{Shaokai Ye}, \bibinfo{person}{Jiayu Li},
  \bibinfo{person}{Jian Tang}, \bibinfo{person}{Wujie Wen},
  \bibinfo{person}{Xue Lin}, \bibinfo{person}{Makan Fardad}, {and}
  \bibinfo{person}{Yanzhi Wang}.} \bibinfo{year}{2018}\natexlab{b}.
\newblock \showarticletitle{Adam-admm: A unified, systematic framework of
  structured weight pruning for dnns}.
\newblock \bibinfo{journal}{\emph{arXiv:1807.11091}} (\bibinfo{year}{2018}).
\newblock


\bibitem[\protect\citeauthoryear{Zhao, Ni, Zhang, Zhao, Zhang, and Tian}{Zhao
  et~al\mbox{.}}{2019}]%
        {zhao2019variational}
\bibfield{author}{\bibinfo{person}{Chenglong Zhao}, \bibinfo{person}{Bingbing
  Ni}, \bibinfo{person}{Jian Zhang}, \bibinfo{person}{Qiwei Zhao},
  \bibinfo{person}{Wenjun Zhang}, {and} \bibinfo{person}{Qi Tian}.}
  \bibinfo{year}{2019}\natexlab{}.
\newblock \showarticletitle{Variational convolutional neural network pruning}.
  In \bibinfo{booktitle}{\emph{Proceedings of the IEEE Conference on Computer
  Vision and Pattern Recognition (CVPR)}}.
\newblock


\bibitem[\protect\citeauthoryear{Zhao, Niu, Yuan, Cai, Sung, Wen, Liu, Shen,
  Ren, Wang, et~al\mbox{.}}{Zhao et~al\mbox{.}}{2020}]%
        {zhao2020achieving}
\bibfield{author}{\bibinfo{person}{Pu Zhao}, \bibinfo{person}{Wei Niu},
  \bibinfo{person}{Geng Yuan}, \bibinfo{person}{Yuxuan Cai},
  \bibinfo{person}{Hsin-Hsuan Sung}, \bibinfo{person}{Wujie Wen},
  \bibinfo{person}{Sijia Liu}, \bibinfo{person}{Xipeng Shen},
  \bibinfo{person}{Bin Ren}, \bibinfo{person}{Yanzhi Wang}, {et~al\mbox{.}}}
  \bibinfo{year}{2020}\natexlab{}.
\newblock \showarticletitle{Achieving Real-Time LiDAR 3D Object Detection on a
  Mobile Device}.
\newblock \bibinfo{journal}{\emph{arXiv preprint arXiv:2012.13801}}
  (\bibinfo{year}{2020}).
\newblock


\bibitem[\protect\citeauthoryear{Zhong, Yan, Wu, Shao, and Liu}{Zhong
  et~al\mbox{.}}{2018}]%
        {zhong2018practical}
\bibfield{author}{\bibinfo{person}{Zhao Zhong}, \bibinfo{person}{Junjie Yan},
  \bibinfo{person}{Wei Wu}, \bibinfo{person}{Jing Shao}, {and}
  \bibinfo{person}{Cheng-Lin Liu}.} \bibinfo{year}{2018}\natexlab{}.
\newblock \showarticletitle{Practical block-wise neural network architecture
  generation}. In \bibinfo{booktitle}{\emph{Proceedings of the IEEE conference
  on computer vision and pattern recognition}}. \bibinfo{pages}{2423--2432}.
\newblock


\bibitem[\protect\citeauthoryear{Zhu, Zhou, and Li}{Zhu et~al\mbox{.}}{2018}]%
        {zhu2018ijcai}
\bibfield{author}{\bibinfo{person}{Xiaotian Zhu}, \bibinfo{person}{Wengang
  Zhou}, {and} \bibinfo{person}{Houqiang Li}.} \bibinfo{year}{2018}\natexlab{}.
\newblock \showarticletitle{Improving Deep Neural Network Sparsity through
  Decorrelation Regularization}. In \bibinfo{booktitle}{\emph{Proceedings of
  International Joint Conferences on Artificial Intelligence (IJCAI)}}.
\newblock


\bibitem[\protect\citeauthoryear{Zhuang, Tan, Zhuang, Liu, Guo, Wu, Huang, and
  Zhu}{Zhuang et~al\mbox{.}}{2018}]%
        {zhuang2018discrimination}
\bibfield{author}{\bibinfo{person}{Zhuangwei Zhuang}, \bibinfo{person}{Mingkui
  Tan}, \bibinfo{person}{Bohan Zhuang}, \bibinfo{person}{Jing Liu},
  \bibinfo{person}{Yong Guo}, \bibinfo{person}{Qingyao Wu},
  \bibinfo{person}{Junzhou Huang}, {and} \bibinfo{person}{Jinhui Zhu}.}
  \bibinfo{year}{2018}\natexlab{}.
\newblock \showarticletitle{Discrimination-aware channel pruning for deep
  neural networks}. In \bibinfo{booktitle}{\emph{Advances in Neural Information
  Processing Systems (NeurIPS)}}.
\newblock


\bibitem[\protect\citeauthoryear{Zoph and Le}{Zoph and Le}{2017}]%
        {zoph2016neural}
\bibfield{author}{\bibinfo{person}{Barret Zoph} {and} \bibinfo{person}{Quoc~V.
  Le}.} \bibinfo{year}{2017}\natexlab{}.
\newblock \showarticletitle{Neural Architecture Search with Reinforcement
  Learning}. In \bibinfo{booktitle}{\emph{International Conference on Learning
  Representations (ICLR)}}.
\newblock


\end{thebibliography}

\appendix

\setcounter{equation}{0}
\setcounter{figure}{0}
\setcounter{table}{0}
\makeatletter
\renewcommand{\theequation}{A\arabic{equation}}
\renewcommand{\thefigure}{A\arabic{figure}}
\renewcommand{\thetable}{A\arabic{table}}


\section{Compiler Optimization Details} \label{app: compiler}
We provide more details of our compiler optimizations in this section. Different from prior DNN inference acceleration frameworks \cite{TensorFlow-Lite,Pytorch-Mobile,Ali-MNN,chen2018tvm,niu2020patdnn,ma2020pconv} that only support dense models or pattern-based pruned models, our compiler optimizations are general, support both dense (unpruned) model and sparse (pruned) model with different pruning schemes for fast inference on various mobile platforms. 
Besides the blocked compressed storage (BCS) and the row reordering optimization mentioned in the main paper (Section~\ref{sec:compiler_opt}),
our compiler-based optimization techniques also include (i) a layer fusion mechanism to fuse different layers together for the reduction of memory consumption of intermediate results and number of operators; 
(ii) an auto-tuning process to determine the best-suited configurations of parameters for different mobile CPUs/GPUs; (iii) Domain Specific Language (DSL) based code generation. 

\subsection{Layer Fusion Mechanism}
To effectively reduce the model inference latency, a layer fusion technique is incorporated in our compiler optimization to fuse the computation operators in the computation graph. With layer fusion, both the memory consumption of the intermediate results and the number of operators can be reduced. The fusion candidates in a model are identified based on two kinds of polynomial calculation properties, i.e., compression laws and data access patterns. The compression laws include associative property, communicative property, and distributive property. 

However, looking for the fusion candidates in such a large space of all combinations of computation operations is too expensive. Therefore, we introduce two constraints to guide the looking up process: (i) only explore the opportunities that are specifically provided due to the above properties, and (ii) only consider enlarging the overall computation for CPU/GPU utilization improvement and reducing the memory access for memory performance improvement as the cost metrics in the fusion. Compared with prior works on loop fusion \cite{ashari2015optimizing,bezanson2017julia,boehm2018optimizing}, our method is more aggressive without high exploration cost.

\subsection{Auto-Tuning for Different Mobile CPUs/GPUs}

During DNN execution, there are many tuning parameters, e.g., matrix tiling sizes, loop unrolling factors, and data placement on GPU memory, that influence the performance. It is hard to determine the best-suited configuration of these parameters manually. To alleviate this problem, our compiler incorporates an auto-tuning approach for both sparse (pruned) model and dense (unpruned) model. The Genetic Algorithm is leveraged to explore the best-suited configurations automatically. It starts parameter search after an initialization with an arbitrary number of chromosomes and explores the parallelism better. Acceleration codes for different DNN models and different mobile CPUs/GPUs can be generated efficiently and quickly through this auto-tuning process, providing the foundation for fast end-to-end inference. The auto-tuning optimizations, together with the layer-fusion and sparse model optimizations, make our framework outperform other acceleration frameworks.

\subsection{DSL-based Code Generation}
In deep learning, a computational graph of a DNN model can be represented by a directed acyclic graph (DAG). Each node in this graph corresponds to an operator.
We propose a high-level Domain Specific Language (DSL) to specify such kind of operators. Each operator in a computational graph also with a layerwise Intermediate Representation (IR) which contains BCS pruning information. The input and output are different tensors in terms of different shapes. This DSL also provides a {\tt Tensor} function for users to create matrices (or tensors).

In this way, DSL is equivalent to a computational graph (that is, DSL is another type of high-level functions used to simulate the data flow of the DNN model), and they can be easily converted to each other. DSL provides users with the flexibility to use existing DNNs or create new DNNs, improving the productivity of DNN programming.
If the DNN already exists, we will convert it into an optimized calculation graph and convert this graph into a DSL. Otherwise, the user writes the model code in the DSL, converts it back to a calculation graph, performs advanced optimization, and regenerates the optimized DSL code.

Finally, our compiler translates the DSL into low-level C++ code for mobile CPU and OpenCL code for mobile GPU, and optimizes the low-level code through a set of optimizations enabled by BCS pruning. The generated code can be then deployed on the mobile device.

\end{document}